\pdfoutput=1  
\documentclass{article}

\usepackage{arxiv}

\usepackage[utf8]{inputenc} 
\usepackage[T1]{fontenc}    
\usepackage[hidelinks]{hyperref} 
\usepackage{url}            
\usepackage{booktabs}       
\usepackage{amsfonts}       
\usepackage{nicefrac}       
\usepackage{microtype}      
\usepackage{natbib}
\usepackage{adjustbox}
\usepackage{verbatim}
\usepackage{multirow}
\usepackage{amsmath,amssymb,amsfonts,pifont}
\usepackage{mathtools}
\usepackage{soul}
\usepackage{blkarray, bigstrut}
\usepackage{amsthm}
\usepackage{lineno}
\usepackage{tikz}
\usepackage{graphicx}
\usepackage{enumitem}
\usepackage{subcaption}
\usepackage{colortbl}
\usepackage{tabularx}
\usetikzlibrary{positioning, calc, shapes.geometric, arrows.meta}
\usepackage{pgf}
\usepackage{centernot}
\usepackage{wrapfig}
\usepackage{changepage}
\usepackage{placeins}
\usepackage{algorithm}
\usepackage{algpseudocode}
\usepackage{needspace}
\usepackage{xcolor}
\definecolor{red}{rgb}{0,0,0}
\usepackage[capitalize,noabbrev]{cleveref}


\theoremstyle{plain}

\newtheorem{proposition}{Proposition}
\newtheorem{lemma}{Lemma}

\theoremstyle{definition}

\theoremstyle{remark}

\theoremstyle{plain}
\newcommand{\restatedname}{}
\newtheorem*{restatedthm}{\restatedname}
\newenvironment{restateprop}[1]
  {\renewcommand{\restatedname}{Proposition~\ref{#1} (restated)}\begin{restatedthm}}
  {\end{restatedthm}}

\title{Estimating Treatment Effects in Networks under Unknown Exposure Mappings}

\author{Daan Caljon\thanks{Corresponding author: \texttt{daan.caljon@kuleuven.be}.}\\
	KU Leuven\\
	\And
	Jente Van Belle\\
	KU Leuven\\
	\And
        Wouter Verbeke\\
        KU Leuven\\
}

\renewcommand{\shorttitle}{Estimating Treatment Effects in Networks under Unknown Exposure Mappings}

\hypersetup{
pdftitle={Estimating Treatment Effects in Networks under Unknown Exposure Mappings},
pdfsubject={cs.LG, stat.ML, stat.ME},
pdfauthor={Daan Caljon, Jente Van Belle, Wouter Verbeke},
pdfkeywords={Causal Inference, Interference, Causal Machine Learning, exposure mapping},
}

\begin{document}
\maketitle

\begin{abstract}
Estimating heterogeneous treatment effects in network settings is complicated by interference, meaning that the outcome of an instance can be influenced by the treatment status of others. Existing causal machine learning approaches that account for interference usually rely on a prespecified exposure mapping 
that summarizes how others' treatments affect the outcome of a given instance, a simplification that is often inappropriate.
We propose HINet, a neural method that combines an expressive GNN outcome model with
network-aware domain-adversarial training. The outcome model learns an exposure-relevant neighborhood representation jointly with outcome prediction, allowing HINet to capture heterogeneous
interference without prespecifying an exposure mapping. HINet's adversarial component uses both node and neighborhood information to promote balance in the learned representations with respect to treatment assignment.
\textcolor{black}{We further derive a population-level generalization bound and introduce two metrics for evaluation across counterfactual networks.}
An empirical evaluation on synthetic and semi-synthetic network datasets demonstrates that HINet performs consistently across diverse exposure mappings, while methods based on a prespecified mapping can perform poorly when it is misspecified.
\end{abstract}



\section{Introduction}
Individualized treatment effect estimation supports data-driven decision-making in applications such as medicine \citep{feuerriegel2024causal}, 
operations management \citep{vanderschueren2023optimizing}, and
economics \citep{varian2016causal}. Traditionally, it is assumed that there is \textit{no interference}, meaning that the treatment assigned to one instance does not affect the outcome of other instances \citep{imbens2015causal,rubin1980randomization}. However, this assumption is often violated in real-world settings due to \textit{spillover effects} \citep{sobel2006randomized,forastiere2021identification}, such as in vaccination, where a vaccine protects not only its recipient but also indirectly benefits their social contacts. \textcolor{black}{If interference is ignored, treatment effect estimates may be biased, and treatment allocation decisions based on these estimates may be suboptimal \citep{forastiere2021identification,caljon2025optimizing}.} 

Recent advances in causal machine learning have introduced methods for estimating treatment effects in networks with interference \citep{ma2021causal,jiang2022estimating,chendoubly}. These methods often rely on a prespecified \emph{exposure mapping}, which encodes how the treatments of other instances in a network influence the outcome of a given instance \citep{aronow2017estimating}. 
A common choice is to define exposure as the sum or proportion of treated one-hop neighbors
\citep{ma2021causal,forastiere2021identification,jiang2022estimating}.
This simplifies estimation and can provide a useful description of interference when the mapping matches the underlying mechanism. In many applications, however, it is unclear how neighbors'
treatments should be combined, as spillover effects may depend on which
neighbors are treated and on their characteristics. A prespecified mapping may then omit relevant information and lead to inaccurate estimates \citep{savje2024misspecified}. 
Estimating treatment effects from observational network data poses a second challenge. Under covariate-dependent treatment assignment, the joint configuration of a node's own treatment and its neighbors' treatments can depend on its local covariate environment (i.e., the covariates of the node and its neighbors). Certain configurations may then be rare for nodes with particular characteristics, requiring the model to extrapolate when estimating counterfactual outcomes. We
refer to this problem as \emph{treatment-configuration imbalance}. Covariate homophily can amplify this problem, because connected nodes tend to have similar covariates and hence similar treatment propensities.

We propose \emph{Heterogeneous Interference Network} (HINet) to estimate treatment effects in networks from observational data.
First, HINet uses an expressive graph neural
network (GNN) to learn an exposure-relevant neighborhood representation
jointly with outcome prediction. This allows HINet's outcome branch to capture heterogeneous spillover effects that
depend on which neighbors are treated and on their characteristics, without requiring a prespecified exposure mapping. Second, HINet uses a network-aware adversarial objective motivated by treatment-configuration imbalance. 
By incorporating neighborhood information into the treatment prediction branch, this objective encourages the learned representations to be less predictive of treatment assignment.
\textbf{Contributions.}\; (1) We propose HINet to estimate heterogeneous treatment effects under network interference without requiring a prespecified exposure mapping.
(2) We motivate its network-aware adversarial objective theoretically and derive a population-level generalization bound relating counterfactual estimation error to factual outcome risk and representation--configuration discrepancy.
(3) We introduce CNEE and PEHNE for evaluating potential-outcome and treatment-effect estimates across counterfactual network assignments.
(4) We then show that HINet performs consistently across multiple datasets, exposure mappings, and strengths of covariate-dependent treatment assignment, whereas methods that rely on a prespecified mapping can be sensitive to its misspecification.

\section{Related Work}

\textbf{Heterogeneous treatment effect estimation.}\;
A large body of work has been dedicated to Conditional Average Treatment Effect (CATE) estimation \citep{hill2011bayesian,shalit2017estimating,kunzel2019metalearners,curth2021nonparametric,vanderschuerenautocate}. 
In contrast to the Average Treatment Effect (ATE), CATEs capture heterogeneity across subpopulations, enabling tailored treatment allocation decisions \citep{feuerriegel2024causal}. 
A key challenge in estimating heterogeneous effects from observational data is the covariate shift
between treated and control units
induced by the treatment assignment mechanism, which can lead to inaccurate treatment effect estimates \citep{johansson2016learning,shalit2017estimating,johansson2022generalization}. 
To address this challenge,
various machine learning methods have been developed, such as propensity weighting and balancing the representations of the treatment and control groups \citep{shalit2017estimating,yao2018representation,hassanpour2019learning}. One notable method closely related to our work is CRN \citep{bica2020estimating}, which employs an adversarial representation balancing approach for estimating treatment effects over time. 

\textbf{Heterogeneous treatment effect estimation under interference.}\; 
Several methods have been developed to estimate treatment effects in network settings with interference.
These methods often rely on a prespecified
exposure mapping that uses a 
basic aggregation function to summarize the neighbors' treatments into a single variable that affects the instance's outcome \citep{aronow2017estimating}. 
Most methods use the proportion of treated one-hop neighbors
as 
the exposure 
mapping.
Under this assumption, 
several 
estimators have been proposed, including inverse probability weighted (IPW) \citep{forastiere2021identification} and doubly robust \citep{chendoubly} estimators. Other methods use GNNs and 
balance the representations of the treatment and control groups by incorporating an Integral Probability Metric (IPM) into the loss function \citep{ma2021causal,cai2023generalization} or by using adversarial training \citep{jiang2022estimating}. Although a prespecified exposure mapping can theoretically be heterogeneous (i.e., dependent on covariates), operationalizing it requires strong assumptions about the interference mechanism. Consequently, most methods based on a prespecified exposure mapping use a homogeneous specification.

Recent work has examined misspecified exposure mappings from different
perspectives: \citet{savje2024misspecified} shows that a mapping can
define meaningful effects without fully capturing the interference
structure, \citet{schroder2026causal} derive
partial-identification bounds under misspecification, and
\citet{huber2026learning} propose a data-driven test of whether a
prespecified mapping captures the relevant interference. Other work focuses on learning exposure mappings directly from data:
IDE-Net \citep{adhikari2025inferring} concatenates multiple expressive candidate mappings and learns a weighted combination approximating the true one,
SPNet \citep{huang2023modeling} leverages masked attention, 
and \citet{zhao2024learning} combine attention weights with Dual Weighted Regression to address covariate shift. 
HINet instead uses an expressive GNN to model how neighborhood treatments and covariates affect outcomes, without prespecifying any exposure mapping, and uses network-aware domain-adversarial training to target the treatment-configuration imbalance induced by covariate-dependent treatment assignment.
A related line of work examines interference over time, i.e., contagion effects,
where outcomes of different entities can also influence each other \citep{jiang2023cf,fatemi2023contagion}. Another strand of work,
which maintains the no-interference assumption,
uses network information to mitigate confounding bias in CATE estimation \citep{guo2020learning,guo2021ignite}.

\section{Problem Setup}\label{sec:Problem setup}
\textbf{Notation.}\;
We consider an undirected network $\mathcal{G}=(\mathcal{V},\mathcal{E})$ where $\mathcal{V}$ is the set of nodes and $\mathcal{E}$ the set of edges connecting the nodes. Each node $i$ is an instance or unit in the network with covariates $\mathbf{X}_i \in \mathcal{X} \subseteq \mathbb{R}^d$, a treatment $T_i \in \mathcal{T} = \{0,1\}$, and an outcome $Y_i \in \mathcal{Y} \subseteq \mathbb{R}$. In marketing, for example, $\mathbf{X}_i$ can represent customer features, $T_i$ whether a customer was targeted with a marketing campaign, and $Y_i$ customer expenditure. The set of directly connected instances, or one-hop neighbors, of instance $i$ is denoted $\mathcal{N}_i$. $\mathcal{N}_i$ is used as a subscript to describe the set of covariates $\mathbf{X}_{\mathcal{N}_i}= \{\mathbf{X}_j\}_{j\in\mathcal{N}_i}$ and treatments $\mathbf{T}_{\mathcal{N}_i}= \{{T}_j\}_{j\in\mathcal{N}_i}$ of $i$'s neighbors.
The potential outcome for instance $i$ with treatment $t_i$ and the set of treatments of its neighbors $\mathbf{t}_{\mathcal{N}_i}$ is denoted as $Y_i(t_i,\mathbf{t}_{\mathcal{N}_i})$. Finally, we denote the local treatment configuration of node $i$ by
$\mathbf{A}_i=(T_i,\mathbf{T}_{\mathcal{N}_i})$ and its local covariate environment by
$\mathbf{C}_i=(\mathbf{X}_i,\mathbf{X}_{\mathcal{N}_i})$.

\begin{wrapfigure}{r}{0.34\textwidth}
    \centering
    \vspace{-6pt}
    \resizebox{0.97\linewidth}{!}{\begin{tikzpicture}[>=stealth, node distance=1.7cm, on grid, auto,
        Xnode/.style={circle,draw,fill = cyan!20},
        Tnode/.style={circle,draw, fill = green!20},
        Ynode/.style={circle,draw, fill = yellow!20},
        every node/.style={minimum size=1cm}
        every path/.style={thick}]  
            \node[Xnode] (Xi) {$\mathbf{X}_i$};
        \node[Xnode, below=1.3cm of Xi] (Xj) {$\mathbf{X}_j$};
        \node[Xnode, above=1.3cm of Xi] (Xk) {$\mathbf{X}_k$};
        
        \node[Tnode, right=of Xi] (Ti) {$T_i$};
        \node[Tnode, below=1.3cm of Ti] (Tj) {$T_j$};
        \node[Tnode, above=1.3cm of Ti] (Tk) {$T_k$};
        
        \node[Ynode, right=of Ti] (Yi) {$Y_i$};
        \node[Ynode, below=1.3cm of Yi] (Yj) {$Y_j$};
        \node[Ynode, above=1.3cm of Yi] (Yk) {$Y_k$};

            \path[->] (Xi) edge (Ti)
                     (Xi) edge (Tj)
                     (Xj) edge (Ti)
                     (Xk) edge[opacity=1] (Tk)  
                     (Xk) edge[opacity=1] (Ti)  
                     (Xi) edge[opacity=1] (Tk)  
                     (Xk) edge[opacity=1] (Tj)  
                     (Xj) edge (Tk)
                     (Xj) edge (Tj)  
                     (Xi) edge[bend left] (Yi)
                     (Xi) edge[opacity=1] (Yj)  
                     (Xi) edge[opacity=1] (Yk)  
                     (Xj) edge (Yi)
                     (Xj) edge[bend right, opacity=1] (Yj)  
                     (Xk) edge[bend left, opacity=1] (Yk)  
                     (Xk) edge[opacity=1] (Yi)  
                     (Ti) edge[opacity=1] (Yk)  
                     (Tk) edge[opacity=1] (Yk)  
                     (Tk) edge (Yi)  
                     (Ti) edge (Yi)
                     (Ti) edge[opacity=1] (Yj)  
                     (Tj) edge (Yi)  
                     (Tj) edge[opacity=1] (Yj);  
\end{tikzpicture}}
    \caption{DAG of the assumed causal structure among three mutually connected network nodes, $i$, $j$, and $k$, each with covariates $\mathbf{X}$, treatment $T$, and outcome $Y$.}
    \label{fig:causal graph}
    \vspace{-6pt}
\end{wrapfigure}

\textbf{Assumptions.}\;
We adopt the Markov assumption: only directly connected instances influence each other. The Directed Acyclic Graph (DAG) of the assumed causal structure is visualized in  \Cref{fig:causal graph} for three mutually connected instances, $i,j,$ and $k$ \citep{greenland1999causal,ogburn2014causal}.
The covariates of unit $i$, $\mathbf{X}_i$, influence the treatment and outcome of both the unit itself and its neighbors. The treatment $T_i$, in turn, affects the outcomes of the unit and its neighbors. 
The arrows from $\mathbf{X}_k$ and $T_k$ to $Y_j$, and from $\mathbf{X}_j$ and $T_j$ to $Y_k$, are omitted for visual clarity. 

We further assume access to an observational dataset $\mathcal{D} = \bigl(\{\mathbf{x}_i,t_i,y_i\}_{i=1}^{|\mathcal{V}|};\mathcal{G}\bigr)$. Importantly, this data does not necessarily come from a randomized controlled trial (RCT), and treatment assignment may depend on covariates, as represented in the DAG by the arrows from a unit's covariates
to its own treatment and the treatments of its neighbors. 
We assume that the treatments and relevant covariates of a node's
neighbors can be summarized by an 
exposure mapping
\citep{aronow2017estimating}, but we do \emph{not} assume that its form is
known. Instead, HINet learns a neighborhood representation jointly with outcome prediction. While this representation is used to predict potential outcomes, it should not be interpreted as recovering the underlying exposure mapping.


Finally, we adapt the classical assumptions from causal inference to ensure identifiability in the interference setting \citep{forastiere2021identification,jiang2022estimating}:

\textit{Consistency:} If $T_i=t_i$ and $\mathbf{T}_{\mathcal{N}_i} = \mathbf{t}_{\mathcal{N}_i},$ then $Y_i = Y_i(t_i,\mathbf{t}_{\mathcal{N}_i})$.

 \textit{Overlap:} 
$\exists \ \delta \in (0,1)$ such that $\delta < p(T_i=t_i,\mathbf{T}_{\mathcal{N}_i}=\mathbf{t}_{\mathcal{N}_i}\mid\mathbf{X}_i=\mathbf{x}_i,\mathbf{X}_{\mathcal{N}_i}=\mathbf{x}_{\mathcal{N}_i}) < 1-\delta$.

\textit{Strong ignorability:} $Y_i(t_i,\mathbf{t}_{\mathcal{N}_i}) \perp\!\!\!\perp T_i,\mathbf{T}_{\mathcal{N}_i} \mid \mathbf{X}_i,\mathbf{X}_{\mathcal{N}_i}$, 
$\forall t_i \in \mathcal{T}, \mathbf{t}_{\mathcal{N}_i}\in \mathcal{T}^{\mathcal{N}_i}, \mathbf{X}_i \in \mathcal{X}, \mathbf{X}_{\mathcal{N}_i} \in \mathcal{X}^{\mathcal{N}_i}$.

\textbf{Objective.}\; We 
aim to 
estimate the Individual Total Treatment Effect (ITTE) \citep{caljon2025optimizing}, defined as: 
\begin{equation} \label{eq:ITTE}
    \omega_i(t_i,\mathbf{t}_{\mathcal{N}_i}) = \mathbb{E}\bigl[Y_i(t_i,\mathbf{t}_{\mathcal{N}_i})-Y_i(0,\mathbf{0})\mid\mathbf{x}_i, \mathbf{x}_{\mathcal{N}_i}\bigr].
\end{equation}
\textcolor{black}{Intuitively, this estimand measures the joint effect of assigning treatment $t_i$ to node $i$ and treatments $\mathbf{t}_{\mathcal{N}_i}$ to its neighbors $\mathcal{N}_i$, compared to not assigning any treatments.}
We train a model $\mathcal{M}(\mathbf{x}_i,t_i,\mathbf{x}_{\mathcal{N}_i},\mathbf{t}_{\mathcal{N}_i})$ to estimate 
$\mathbb{E}\bigl[Y_i(t_i,\mathbf{t}_{\mathcal{N}_i})\mid \mathbf{x}_i,\mathbf{x}_{\mathcal{N}_i}\bigr]$, which is then plugged into the definition of the ITTE to obtain $\hat{\omega}_i(t_i,\mathbf{t}_{\mathcal{N}_i})=\mathcal{M}(\mathbf{x}_i,t_i,\mathbf{x}_{\mathcal{N}_i},\mathbf{t}_{\mathcal{N}_i})-\mathcal{M}(\mathbf{x}_i,0,\mathbf{x}_{\mathcal{N}_i},\mathbf{0})$. 

\textbf{Measuring model performance under interference.}\;
In a traditional no-interference setting with binary treatment,
the Precision in Estimation of Heterogeneous Effects (PEHE) \citep{hill2011bayesian,shalit2017estimating} is often used to evaluate treatment effect estimates on \mbox{(semi-)synthetic} data. PEHE is defined as the mean squared error between estimated and ground-truth CATEs. In a binary treatment setting, there is a single treatment effect. In a network setting, this is no longer the case as there are \textit{many possible treatment assignments}, each resulting in different potential outcomes. 
Previous work has typically evaluated performance based on only one \textit{counterfactual network}, i.e., a network in which at least one node receives a different treatment than observed (see, e.g., \citet{jiang2022estimating,chendoubly}). 
However, some models may be accurate for certain counterfactual networks (e.g., those with a low treatment rate) but perform poorly for others. Therefore, we argue that a comprehensive evaluation procedure should
account for \textit{performance across a range of counterfactual networks}.

To address this issue,
we propose two {evaluation metrics} inspired by the Mean Integrated Squared Error (MISE) \citep{schwab2020learning}, which is used for evaluating continuous treatment effect estimates:
the \emph{Precision in Estimation of Heterogeneous Network Effects} (PEHNE) and the \emph{Counterfactual Network Estimation Error} (CNEE).
Both are defined as the expected squared estimation error over a target distribution of counterfactual networks: 
PEHNE in terms of ITTE and CNEE in terms of counterfactual outcomes. The key difference is that CNEE places less emphasis on the estimation of potential outcomes without any treatment, $Y_i(0,\mathbf{0})$.
Since there are $2^{|\mathcal{V}|}-1$ possible counterfactual networks, it is computationally infeasible to evaluate all for large networks. 
Therefore, we sample $m$ counterfactual networks, calculate the squared estimation error for each node within each sampled counterfactual network, and average across nodes and counterfactual networks. 
In our experiments, we sample $m=50$ counterfactual networks, with treatment rates spanning zero to one, and treated nodes sampled uniformly at each rate. Further details, including pseudocode, are provided in \Cref{app: Performance metrics}.
\textbf{Treatment-configuration imbalance.}\;
In observational data, treatment assignment may depend on covariates through a policy or self-selection, inducing covariate shift between treated and
control units, a well-known source of treatment-effect estimation error
\cite{shalit2017estimating,johansson2022generalization}. Under interference,
however, the relevant treatment assignment for node $i$ is its local treatment
configuration $\mathbf{A}_i$,
which includes both its own treatment and those of its
neighbors. Similarly, the relevant covariates are its local
covariate environment $\mathbf{C}_i$.
When treatment assignment depends on covariates, the distribution of $\mathbf{A}_i$ can vary with $\mathbf{C}_i$. Even when overlap holds, some treatment configurations may be infrequently observed for particular local covariate environments, requiring greater extrapolation when estimating counterfactual outcomes. We refer to this network analogue of covariate imbalance across treatment groups as \textit{treatment-configuration imbalance}.

Homophily can amplify this imbalance. When connected nodes have similar
covariates and treatment assignment depends on those covariates, connected
nodes also tend to have similar treatment propensities. Treated and untreated
nodes can therefore form local clusters, strengthening the association
between covariates and treatment configurations. Importantly, under our assumption that the relevant covariates are observed, homophily does not by itself invalidate identification: strong ignorability conditional on the local covariate environment remains sufficient \citep{shalizi2011homophily}. \Cref{app:homophily_dag} provides the causal structure under homophily. Treatment-configuration imbalance motivates HINet's network-aware
domain-adversarial training, introduced in the next section. 

\section{Methodology}
\begin{wrapfigure}{r}{0.52\textwidth}
    \centering
    \vspace{-6pt}
    \resizebox{0.98\linewidth}{!}{\resizebox{0.45\textwidth}{!}{\begin{tikzpicture}[x=2cm, y=1.5cm, >=Stealth]

\tikzstyle{inputnode} = [circle, draw, minimum size=1cm, fill=cyan!0]
\tikzstyle{gnnnode} = [rectangle, draw, rounded corners, minimum size=1cm, fill=gray!20]
\tikzstyle{hiddennode} = [rectangle, draw, rounded corners, minimum width=0.3cm, minimum height=1cm, fill=gray!20]
\tikzstyle{outputnode} = [circle, draw, minimum size=1cm, fill=yellow!0]

\tikzstyle{invis} = [circle, draw, minimum size=1cm, fill=yellow!0,]
\tikzstyle{discriminator} = [circle, draw=none, fill=none]
\tikzstyle{GRL} = [rectangle, draw, rounded corners, minimum width=0.3cm, minimum height=1cm, fill=gray!20]
\node[inputnode] (xi) at (0,0) {$\mathbf{x}$};

\node[hiddennode, right=0.5cm of xi] (hidden) {};


\node[outputnode, right=0.7cm of hidden] (phi) {$\boldsymbol{\phi}$};

\coordinate (phiup) at ($(phi.north) + (0,0.95cm)$);
\coordinate (phidown) at ($(phi.south) - (0,0.7cm)$);


\node[gnnnode,right= 1cm of phiup,rotate=0] (TGIN) {$\text{GIN}_T$};
\coordinate (TGINright) at ($(TGIN.east) +(1.4cm,0cm)$);

\node[hiddennode,right=0.35cm of phidown]
  (messageY) {};

\node[discriminator,above=-0.2cm of messageY]
  (message_encoder) {$e_m$};

\node[gnnnode,right=0.40cm of messageY,rotate=0]
  (YGIN) {$\text{GIN}_Y$};
\node[GRL,left=0.4cm of TGIN,rotate=0] (GRLGIN) {\parbox{0.2cm}{\centering G\vspace{-2.5pt}\\R\vspace{-2.5pt}\\L}};
\node[hiddennode, below= -0.22cm of TGINright] (hiddenT){};
\node[hiddennode, right= 1.25cm of YGIN] (hiddenY){};
\node[inputnode, below=0.2cm of hidden] (t) {$\mathbf{t}$};

\node[discriminator, above= -0.2cm of hidden] (encoder){$e_{\phi}$};
\node[discriminator, above= -0.2cm of hiddenT] (d_t){$d_T$};
\node[discriminator, above= -0.2cm of hiddenY] (p_Y){$p_Y$};

\node[outputnode, right=0.5cm of hiddenT] (That){$\hat{{t}}$};
\node[outputnode, right=0.5cm of hiddenY] (Yhat){$\hat{{y}}$};

\coordinate (middle) at ($(phi.east) + (1.8cm,0)$);

\coordinate (hiddenTbelow) at ($(hiddenT.west) - (0,0.3cm)$);
\coordinate (hiddenTabove) at ($(hiddenT.west) + (0,0.3cm)$);
\coordinate (GINright) at ($(middle) + (0,1.1cm)$);
\coordinate (hiddenYbelow) at ($(hiddenY.west) - (0,0.3cm)$);
\coordinate (hiddenYabove) at ($(hiddenY.west) + (0,0.3cm)$);
\coordinate (YGINabove) at ($(YGIN.west) + (0,0.3cm)$);
\coordinate (t_helper) at ($(YGIN.south) - (0,0.3cm)$);
\coordinate (py_left) at ($(hiddenYbelow) - (0.3,0.0cm)$);
\coordinate (messageYabove)
  at ($(messageY.west) + (0,0.25cm)$);

\coordinate (messageYbelow)
  at ($(messageY.west) - (0,0.25cm)$);
\node[GRL,left=0.4cm of hiddenTbelow,rotate=0] (GRLIND) {\parbox{0.2cm}{\centering G\vspace{-2.5pt}\\R\vspace{-2.5pt}\\L}};






\draw[->] (xi) -- (hidden);

\draw[->] (hidden) -- (phi);
\draw[->] (phi) |- (GRLGIN) -- (TGIN);

\draw[->] (phi) |- (messageYabove);
\draw[->] (t) -- (messageY);
\draw[->] (messageY) -- (YGIN);



\draw[->] (TGIN) --(hiddenTabove);
\draw[->] (YGIN) -- (hiddenY);

\draw[->] (phi) -- (middle) |- (hiddenYabove);
\draw[->] (phi) -- (middle) |- (GRLIND) --(hiddenTbelow);

\draw[->] (hiddenY) -- (Yhat);
\draw[->] (hiddenT) -- (That);
\draw[->] (t) |- (t_helper) -| (py_left) -- (hiddenYbelow);

\end{tikzpicture}}}
   \caption{HINet architecture: the lower branch performs outcome prediction, while the upper branch handles representation balancing using Gradient Reversal Layers (GRLs).}
    \label{fig:model}
    \vspace{-6pt}
\end{wrapfigure}
\textbf{Architecture.}\;
\Cref{fig:model} visualizes HINet.
Inspired by \citet{bica2020estimating} and \citet{berrevoets2020}, 
HINet first learns node representations shared by two branches: a lower branch predicting outcomes and an upper branch predicting treatment and providing the adversarial training signal. For each node $k \in \{i\} \cup \mathcal{N}_i$, the covariates $\mathbf{x}_k$ are transformed into a representation 
$\boldsymbol{\phi}_k= e_{\phi}(\mathbf{x}_k)$
via a multi-layer perceptron (MLP),
and the resulting node representations are then used by the two branches to predict 
the outcome $\hat{y}_i$
and the treatment $\hat{t}_i$ of node $i$, respectively. 

The \textit{lower branch} predicts the outcome of node $i$.
To model how neighborhood treatments and covariates affect the outcome
without specifying an exposure mapping, we use an expressive GNN. In this paper, we instantiate it using the Graph Isomorphism Network (GIN) \citep{xu2018powerful}. 
This module takes as input the node representations $\boldsymbol{\phi}_k$ and treatments $t_k$ for $k \in \{i\} \cup \mathcal{N}_i$.
We choose GIN for its theoretical properties: its aggregation retains information about neighborhood size and can represent count-based exposure mappings \citep{xu2018powerful}. This makes it a suitable default when the functional form of the exposure mapping is unknown. 
Although we use GIN throughout, HINet can accommodate other message-passing GNNs, such as GraphSAGE \citep{hamilton2017inductive}.
The lower branch first concatenates the node representation and treatment
of each node $k\in\{i\}\cup\mathcal{N}_i$, i.e.,
$\mathbf{q}_k=\boldsymbol{\phi}_k\oplus t_k$.
An MLP $e_m$ is then applied to obtain
$\mathbf{m}_k=e_m(\mathbf{q}_k)$. This transformation allows
the GIN inputs to depend on both a neighbor's representation and its treatment.
$\mathrm{GIN}_Y$ subsequently outputs
\[    \mathbf{h}^{Y}_i
    =
    \mathrm{MLP}_Y\bigl(
        (1+\epsilon)\mathbf{m}_i
        +
        \sum_{j\in\mathcal{N}_i}\mathbf{m}_j\bigr)
,
\]
where $\epsilon$ controls the relative weight assigned to node $i$'s representation. The resulting neighborhood representation
$\mathbf{h}^{Y}_i$ is combined with $\boldsymbol{\phi}_i$ and $t_i$, and passed to the MLP
$p_Y$ to generate the outcome prediction $\hat{y}_i$, preserving a direct path from node $i$'s representation and treatment to its outcome.


The \textit{upper branch} is used to induce treatment-invariant, or balanced, representations $\boldsymbol{\phi}$. Like the lower branch, it processes the representations of node $i$ and its neighbors using a GIN, but treatments are not used as inputs. The resulting representation 
and $\boldsymbol{\phi}_i$ are used by the MLP $d_T$ to predict the treatment $t_i$ of node $i$. Gradient Reversal Layers (GRLs) \citep{ganin2016domain}, which do not affect the forward pass but reverse the gradient from treatment prediction during the backward pass, are used to induce the encoder $e_\phi$ to produce representations from which treatment is less predictable.

\textbf{Loss function.}\;
HINet is trained by combining two different losses: the outcome loss and the treatment prediction loss, defined respectively as $\mathcal{L}_y = \frac{1}{n}\sum_{i=1}^n(y_i-\hat{y}_i)^2$ and $\mathcal{L}_t=\frac{1}{n}\sum_{i=1}^n\text{BCE}(t_i,\hat{t}_i)$, where $n$ denotes the number of training nodes and $\text{BCE}$ denotes the binary cross-entropy loss.
The two losses are combined as
\begin{equation}
    \mathcal{L}_{\text{comb}}=\mathcal{L}_y + \alpha\cdot\mathcal{L}_t,
\end{equation}
where $\alpha$ controls the strength of adversarial balancing. 

Each branch is updated only by its own loss: $\mathcal{L}_y$ updates the lower branch, $\mathcal{L}_t$ the upper branch, and both update the shared encoder $e_\phi$. The upper branch is optimized to minimize $\mathcal{L}_t$, whereas $e_\phi$ is updated through the GRLs to increase it.

\textbf{Representation balancing.}\;
Intuitively, balancing aims to reduce the dependence between the learned representations and treatment assignment \citep{shalit2017estimating,johansson2022generalization}. 
Because the upper branch predicts the treatment of node $i$ using the representations of both the node and its neighbors, the balancing objective targets the local representation $\boldsymbol{\Phi}_i=(\boldsymbol{\phi}_i,\boldsymbol{\phi}_{\mathcal{N}_i})$. Under exact balance, this objective corresponds to
\begin{equation}
\label{eq:treatment-invariance}
p(\boldsymbol{\Phi}_i\mid T_i) = p(\boldsymbol{\Phi}_i) \quad \forall i \in \mathcal{V}.
\end{equation}

\begin{proposition}\label{theorem:neighbors}
Assume that \Cref{eq:treatment-invariance} holds for every node. Under the conditional independence conditions implied by the causal structure in \Cref{fig:causal graph}, the representations are also invariant with respect to each neighbor's treatment:
\[
p(\boldsymbol{\Phi}_i\mid T_j)
=
p(\boldsymbol{\Phi}_i),
\qquad
\forall i\in\mathcal{V},\ \forall j\in\mathcal{N}_i.
\]
\end{proposition}

The proof is provided in \Cref{app:proof_prop1}. \Cref{theorem:neighbors} considers an idealized setting in which exact invariance holds at every node. Under the implied conditional independence conditions, invariance of the local representations with respect to each node's own treatment also implies marginal invariance with respect to the treatments of its neighbors. In practice, domain-adversarial training can only approximate this invariance. Nevertheless, this result motivates incorporating neighborhood information into HINet's treatment-prediction branch. Under homophily, the representations must additionally retain the covariate information needed to block the dependence induced by conditioning on the neighborhood. \Cref{app: homophily interference} discusses this requirement.

\textbf{Generalization bound.}\;
We now relate counterfactual estimation error to factual outcome prediction and representation--configuration discrepancy:
\begin{proposition}\label{prop:bound}
Under the identification assumptions introduced above, a representation-sufficiency condition requiring the conditional outcome loss to depend on $\mathbf{C}_i$ only through $\boldsymbol{\Phi}_i$, and standard IPM regularity, the counterfactual-outcome error is upper-bounded as follows:
\begin{equation*}
    \epsilon_{\mathrm{CNEE}}(\mathcal{M}) \le R^{F}(\mathcal{M}) + C_{\boldsymbol{\Phi}}\,\mathrm{IPM}_{\mathcal{L}}\big(p^{\rho}_{\boldsymbol{\Phi}},\,p^{\mathrm{obs}}_{\boldsymbol{\Phi}}\big) - \sigma^2.
\end{equation*}
\end{proposition}

In this inequality, $R^{F}$ is the factual outcome risk, while $\mathrm{IPM}_{\mathcal{L}}$ is an integral probability metric between $p^{\mathrm{obs}}_{\boldsymbol{\Phi}}$, the observed joint distribution $(\boldsymbol{\Phi}_i,\mathbf{A}_i)$, and $p^{\rho}_{\boldsymbol{\Phi}}$, the target distribution induced by $\rho$, which specifies how the counterfactual assignments used to evaluate CNEE and PEHNE are sampled. Furthermore, $C_{\boldsymbol{\Phi}}$ is a constant and $\sigma^2$ denotes irreducible outcome noise. $\epsilon_{\mathrm{PEHNE}}$ has an analogous bound with an additional discrepancy term for the zero treatment configuration. Both bounds adapt the bounds of \citet{johansson2022generalization} to the interference setting. Unlike the network bound of \citet{cai2023generalization}, which assumes a known exposure mapping, this bound is stated at the level of local treatment configurations. The complete proofs and bounds are in \Cref{app:bound}.

\Cref{prop:bound} separates counterfactual estimation error into factual
outcome risk and a representation--configuration discrepancy. This decomposition motivates the two components of HINet's objective. The outcome loss
$\mathcal{L}_y$ estimates the factual risk, while the adversarial loss
$\mathcal{L}_t$ makes treatment less predictable from the learned
representations and therefore targets one source of dependence
that contributes to this discrepancy. Under \Cref{theorem:neighbors}, invariance of the learned representations with respect to a node's own treatment also implies marginal invariance with respect to each neighbor's treatment.
Nevertheless, HINet does not optimize the configuration-level IPM. Directly balancing $\mathbf{A}_i$ is impractical because, for a node with $|\mathcal{N}_i|$ neighbors, it can take up to $2^{|\mathcal{N}_i|+1}$ values. Moreover, each node is observed under only one configuration, and neighborhood sizes vary. Reweighting faces the same problem through the joint propensity $p(\mathbf{A}_i\mid\mathbf{C}_i)$, while summarizing $\mathbf{A}_i$ before balancing would require a prespecified exposure mapping. HINet therefore uses \emph{marginal adversarial balancing} as a tractable alternative to balancing with respect to the joint treatment configuration. This can reduce dependence on the individual treatments in the configuration, although dependence through \emph{joint} treatment patterns may remain. Increasing $\alpha$ places greater emphasis on marginal balancing but may increase the factual risk and induce representation-induced confounding bias if pushed too far \citep{melnychuk2024bounds}. Our experiments examine this trade-off empirically.

\section{Experiments and Discussion}\label{sec:experiments}
\textbf{Data.}\;
Synthetic and semi-synthetic data are commonly used in causal machine learning to evaluate treatment effect estimators, as ground truth effects are unobservable in real-world datasets \citep{feuerriegel2024causal}. We use two fully synthetic datasets: one using the \textit{Barabási–Albert (BA Sim)} random network model \citep{barabasi1999emergence}, 
and another using a procedure that generates {homophilous} \textit{(Homophily Sim)} graphs based on cosine similarity. Following related work \citep{ma2021causal,jiang2022estimating,chendoubly}, we additionally construct semi-synthetic datasets from \textit{Flickr} and \textit{BlogCatalog (BC)} networks.
We also include Coauthor-CS \cite{shchur2019pitfallsgraphneuralnetwork}, a larger real-world graph with homophilous node features. For these three semi-synthetic datasets, the network structures and node features are observed, while treatments and outcomes are simulated so that all counterfactual outcomes are available.

Each dataset has train/validation/test splits: 10,000 nodes per split for BA Sim and Homophily Sim, $\approx$2,400 for Flickr, $\approx$1,700 for BC, and 6,111 for Coauthor-CS.
Unless explicitly stated otherwise, we use a feature-weighted mean as the default exposure mapping in the data-generating process (DGP), i.e., $z_i= \frac{1}{\mid \mathcal{N}_i\mid} \sum_{j\in \mathcal{N}_i} w(\mathbf{x}_j) t_j$, with $w(\cdot)$ a function that maps node features to a weight. 
This exposure mapping is used only to generate the outcomes. Its form is \emph{not} provided to any method during training.
Full details on the DGPs are provided in \Cref{app: DGP}.


\begin{table}[t]
\centering
\footnotesize
\begin{adjustbox}{width=\textwidth}
\captionsetup{width=\textwidth}
\begin{tabular}{lccccccccc}
\toprule
Dataset & Metric & TARNet & NetDeconf & NetEst & TNet & GIN model & SPNet & IDE-Net & HINet (ours) \\
\midrule
\multirow{2}{*}{BC} & PEHNE & 1.53 $\pm$ 0.29 & 5.28 $\pm$ 0.13 & 2.18 $\pm$ 0.22 & 1.71 $\pm$ 0.32 & \underline{1.13 $\pm$ 0.09} & 5.08 $\pm$ 0.13 & 1.48 $\pm$ 0.33 & \textbf{0.53 $\pm$ 0.13} \\
 & CNEE & 2.42 $\pm$ 0.27 & 5.26 $\pm$ 0.11 & 2.12 $\pm$ 0.17 & 1.82 $\pm$ 0.31 & \underline{0.92 $\pm$ 0.11} & 4.93 $\pm$ 0.12 & 1.49 $\pm$ 0.32 & \textbf{0.58 $\pm$ 0.14} \\
\midrule
\multirow{2}{*}{Flickr} & PEHNE & \underline{0.93 $\pm$ 0.26} & 6.14 $\pm$ 0.26 & 3.18 $\pm$ 0.21 & 1.42 $\pm$ 0.07 & 1.12 $\pm$ 0.08 & 6.80 $\pm$ 0.33 & 2.18 $\pm$ 0.42 & \textbf{0.71 $\pm$ 0.21} \\
 & CNEE & 2.84 $\pm$ 0.29 & 7.68 $\pm$ 0.20 & 4.67 $\pm$ 0.26 & 2.52 $\pm$ 0.07 & \underline{1.19 $\pm$ 0.11} & 8.26 $\pm$ 0.42 & 2.55 $\pm$ 0.47 & \textbf{0.78 $\pm$ 0.23} \\
\midrule
\multirow{2}{*}{BA Sim} & PEHNE & 0.80 $\pm$ 0.10 & 4.54 $\pm$ 0.01 & 0.93 $\pm$ 0.12 & 0.83 $\pm$ 0.06 & 1.69 $\pm$ 0.13 & 4.97 $\pm$ 0.33 & \underline{0.65 $\pm$ 0.08} & \textbf{0.28 $\pm$ 0.03} \\
 & CNEE & 3.55 $\pm$ 0.08 & 6.51 $\pm$ 0.01 & 1.80 $\pm$ 0.10 & 1.42 $\pm$ 0.07 & 1.49 $\pm$ 0.12 & 6.71 $\pm$ 0.30 & \underline{0.65 $\pm$ 0.07} & \textbf{0.35 $\pm$ 0.07} \\
\midrule
\multirow{2}{*}{Homophily Sim} & PEHNE & 1.04 $\pm$ 0.06 & 1.00 $\pm$ 0.03 & 0.58 $\pm$ 0.08 & 0.83 $\pm$ 0.06 & \underline{0.51 $\pm$ 0.03} & 1.77 $\pm$ 0.04 & 0.64 $\pm$ 0.08 & \textbf{0.17 $\pm$ 0.07} \\
 & CNEE & 1.49 $\pm$ 0.06 & 1.08 $\pm$ 0.01 & 0.81 $\pm$ 0.07 & 0.88 $\pm$ 0.05 & \underline{0.66 $\pm$ 0.03} & 1.74 $\pm$ 0.04 & 0.68 $\pm$ 0.07 & \textbf{0.20 $\pm$ 0.06} \\
 \midrule
\multirow{2}{*}{Coauthor-CS} & PEHNE & \underline{1.18 $\pm$ 0.15} & 2.94 $\pm$ 0.77 & 3.08 $\pm$ 0.23 & 1.41 $\pm$ 0.15 & 2.19 $\pm$ 0.15 & 4.00 $\pm$ 2.37 & 1.84 $\pm$ 0.39 & \textbf{1.14 $\pm$ 0.23} \\
 & CNEE & 2.41 $\pm$ 0.08 & 2.83 $\pm$ 0.44 & 2.91 $\pm$ 0.06 & \underline{1.75 $\pm$ 0.13} & 3.10 $\pm$ 0.40 & 3.59 $\pm$ 0.92 & 2.07 $\pm$ 0.34 & \textbf{1.37 $\pm$ 0.34} \\
\bottomrule
\end{tabular}
\end{adjustbox}
\caption{Test set results (mean $\pm$ SD over five different initializations). Lower is better for both
metrics. The best-performing method is in bold; the second-best is underlined.}
\label{tab:results}
\end{table}

\begin{figure}[t]
  \centering
  \begin{subfigure}[b]{.19\textwidth}
  \includegraphics[width=\textwidth]{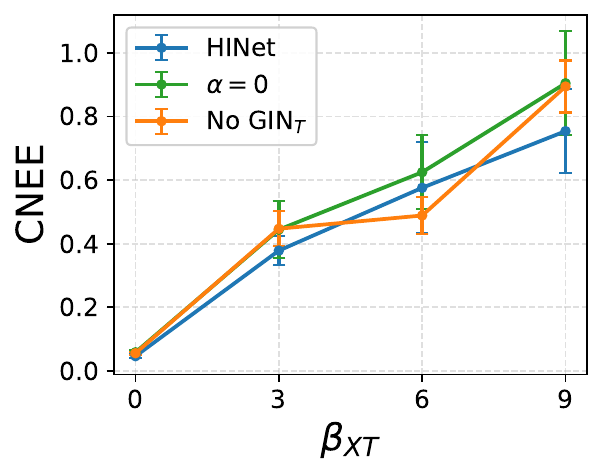}
    \caption{BC}
  \end{subfigure}
  \begin{subfigure}[b]{.19\textwidth}
    \includegraphics[width=\textwidth]{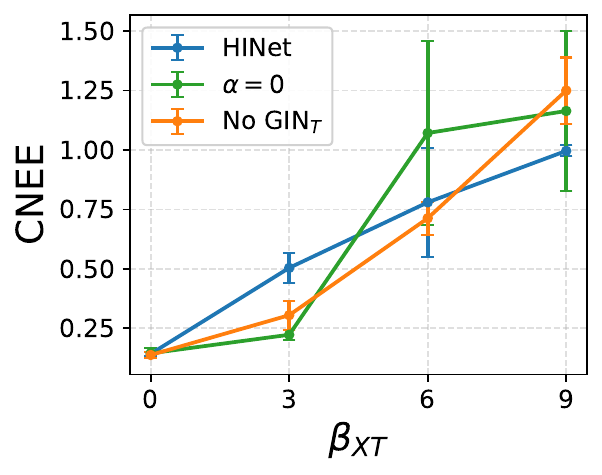}
    \caption{Flickr}
  \end{subfigure}
  \begin{subfigure}[b]{.19\textwidth}
    \includegraphics[width=\textwidth]{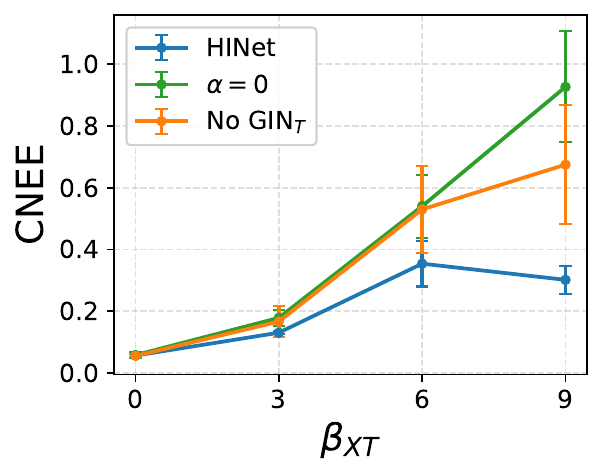}
    \caption{BA Sim}
  \end{subfigure}
  \begin{subfigure}[b]{.19\textwidth}
    \includegraphics[width=\textwidth]{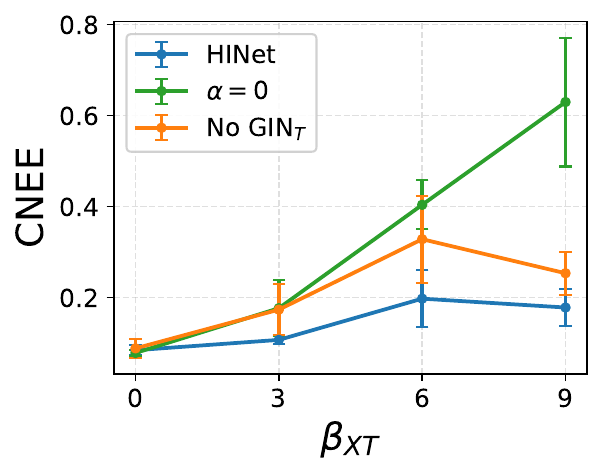}
    \caption{Homophily Sim}
  \end{subfigure}
  \begin{subfigure}[b]{.19\textwidth}
    \includegraphics[width=\textwidth]{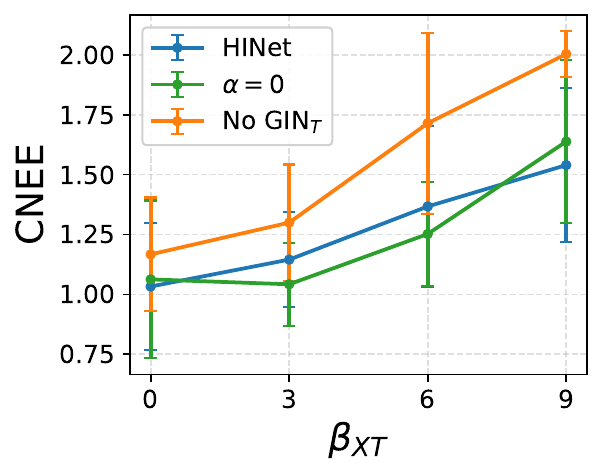}
    \caption{Coauthor-CS}
  \end{subfigure}

  \caption{CNEE across datasets with increasing covariate-dependent treatment assignment strength $\beta_{XT}$ (mean $\pm$ SD over five different initializations). Lower is better.}
  \label{fig:cnee_rep_bal}
\end{figure}

\textbf{Methods for comparison.}\;
We compare HINet with the following 
methods for estimating treatment effects: 
\textit{TARNet} \citep{shalit2017estimating}, which ignores network information; 
\textit{NetDeconf} \citep{guo2020learning}, which incorporates network information but does not account for spillover effects; 
\textit{NetEst} \citep{jiang2022estimating}, which relies on a prespecified exposure mapping to estimate spillover effects; \textit{TNet} \citep{chendoubly}, which also relies on a prespecified exposure mapping, but leverages targeted learning \citep{vanderLaan_targeted} for doubly robust estimation of spillover effects; our implementation of \textit{SPNet} \citep{huang2023modeling}, which 
estimates heterogeneous spillover effects using a masked attention mechanism; and
\textit{IDE-Net} \citep{adhikari2025inferring}, which constructs and concatenates multiple candidate exposure mappings, allowing the model to learn from several neighborhood-treatment summaries rather than relying on a single prespecified mapping.
Finally, 
we include a \textit{GIN model}, which uses node features and treatments as inputs to a GIN layer that is followed by an MLP.
It is a naive network-aware baseline, not an architecture-matched ablation of HINet. The $\alpha=0$ HINet variant in \Cref{fig:cnee_rep_bal} isolates the contribution of adversarial balancing. For NetEst and TNet, which require a prespecified exposure mapping, we follow the literature and use
 $z_i^{\text{assumed}}=\frac{1}{\mid \mathcal{N}_i\mid}\sum_{j\in \mathcal{N}_i}  t_j$.
This mapping is correctly specified when the DGP uses the proportion of treated neighbors as the exposure mapping and misspecified under the other exposure mappings, allowing us to evaluate sensitivity to exposure-mapping misspecification.

\begin{figure}
    \centering
    \includegraphics[width=0.98\textwidth]{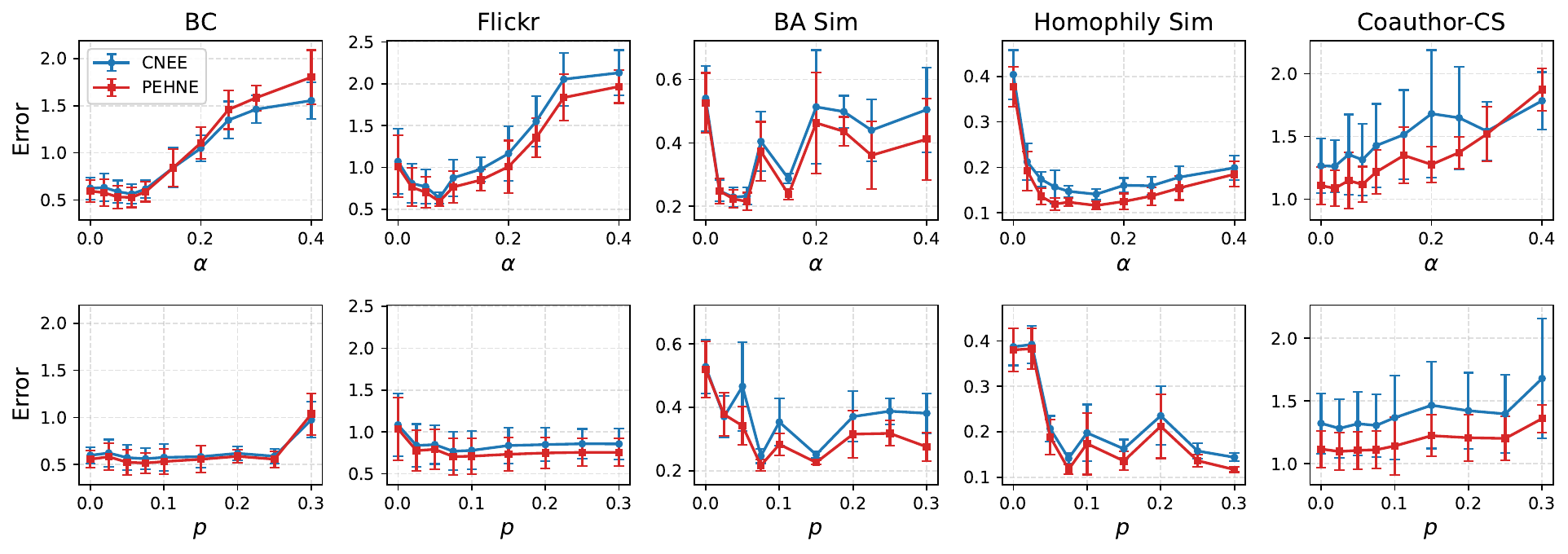}
    \caption{Sensitivity of HINet's test CNEE and PEHNE (mean $\pm$ SD over five different initializations) to the balancing weight $\alpha$ (top row) and the selection tolerance $p$ (bottom row). Columns correspond to the five datasets.  Lower is better for both metrics.}
      \label{fig:sensitivity}
\end{figure}

\textbf{Hyperparameter selection.}\;
Evaluation metrics based on counterfactual outcomes, such as PEHNE and CNEE, are available in our simulated settings but not in real applications. Hence, we do not use them for model selection. Instead, we first tune the standard hyperparameters of all methods using the \emph{factual validation loss}, i.e., the prediction error on the observed validation set outcomes \citep{curth2023search}.
Because balancing can trade factual predictive accuracy for treatment invariance, this loss generally favors $\alpha=0$.
Nevertheless, \Cref{prop:bound} and prior work on balancing representations \citep{shalit2017estimating,bica2020estimating} motivate the use of positive values of $\alpha$, but selecting this parameter remains challenging when a suitable observable proxy for counterfactual error is unavailable. We therefore use a selection heuristic based only on factual validation outcomes.
After fixing the standard hyperparameters, we select $\alpha$ as the largest value for which the factual validation loss does not exceed $(1+p)\cdot\text{loss}_{\alpha=0}$. 
In our experiments, we fix $p=0.10$, allowing for a maximum validation error increase of 10\% relative to $\alpha=0$. 
Because $\alpha=0$ is a candidate, the heuristic defaults to no balancing when none of the positive candidate values satisfies this constraint. 
To ensure a consistent comparison, we use the same procedure for NetEst's balancing weight.
Details on hyperparameter selection for all methods are provided in \Cref{app:Hyperparam selection}. Sensitivity to $\alpha$ and $p$ is shown in \Cref{fig:sensitivity}.

\textbf{Performance under the default exposure mapping.}\;
\Cref{tab:results} reports the test set results. 
HINet achieves the lowest PEHNE and CNEE on all five datasets. The improvements are largest on BC, BA Sim, and Homophily Sim, whereas the
differences on Flickr and Coauthor-CS are smaller relative to the variation
across initializations. These results provide initial evidence that HINet can estimate treatment effects under interference across different network structures without prespecifying an exposure mapping. 


\begin{table}[t]
    \centering
    \small
    \setlength{\tabcolsep}{14pt}
    \captionsetup{width=0.66\textwidth}
    \begin{tabular}{lcccc}
        \toprule
        Method
        & CNEE
        & PEHNE
        & Best
        & Top-2 \\
        \midrule
        TARNet        & 6.0 & 5.4 &  0 &  2 \\
        NetDeconf     & 6.7 & 7.0 &  0 &  0 \\
        NetEst        & 5.2 & 4.8 &  0 &  2 \\
        TNet          & 4.1 & 3.7 &  5 & 13 \\
        GIN model     & 3.0 & 3.1 &  5 & 27 \\
        SPNet         & 6.8 & 7.2 &  0 &  0 \\
        IDE-Net       & 3.2 & 3.6 &  0 &  6 \\
        \textbf{HINet}
                      & \textbf{1.1}
                      & \textbf{1.3}
                      & \textbf{40}
                      & \textbf{50} \\
        \bottomrule
    \end{tabular}
        \caption{%
        Average rank over 25 dataset--mapping combinations, along with the
        number of metric comparisons in which each method achieves the best or top-2 mean performance.
    }
    \label{tab:mapping_summary}
\end{table}

\textbf{Performance under other exposure mappings.}\;
We repeat the experiments from \Cref{tab:results} using four alternative mappings in the DGP: the sum of neighbors' treatments, the
proportion of treated neighbors, information entropy of the
neighborhood treatment rate, and a feature-weighted mean with squared weights.
\Cref{tab:mapping_summary} summarizes performance across all exposure
mappings (default and four alternatives) and datasets.
Detailed results are provided in
\Cref{app:other_mappings}.
Across the 25 dataset--mapping combinations and the two evaluation metrics,
HINet performs best in 40 out of 50 comparisons and second-best in the remaining ten. The best
competing method depends on the mapping used in the DGP. On Coauthor-CS---the largest
real-world network considered---HINet achieves the lowest CNEE under
every mapping and the best average rank for both metrics. The relative performance of methods that use a prespecified mapping depends on how closely this assumed mapping approximates the true mechanism. NetEst and TNet are substantially less accurate than HINet under the sum mapping on every dataset, whereas TNet remains competitive for mean-like mappings, which its assumed proportion mapping reasonably approximates. HINet's consistently good performance across mappings is relevant in practice, as the true mechanism is generally unknown and difficult to assess from factual data.

\textbf{Representation balancing.}\;
Using the default exposure mapping, we
increase the strength of covariate-dependent treatment assignment $\beta_{XT}$ and compare HINet with variants without balancing ($\alpha=0$) and without $\mathrm{GIN}_T$. The latter accounts only for each node's own representation during balancing,
and thus does not target the invariance condition in \Cref{eq:treatment-invariance}, so \Cref{theorem:neighbors} does not apply. \Cref{fig:cnee_rep_bal} reports CNEE across
all five datasets.
Under covariate-independent treatment assignment, the three
variants perform similarly. At moderate $\beta_{XT}$, the
effect of balancing varies by dataset. At the largest $\beta_{XT}$ values, however, HINet has the lowest mean CNEE on all
datasets, with the largest improvements over the $\alpha=0$ variant 
on BA Sim and Homophily Sim. 
The comparison with No GIN$_T$ shows the importance of incorporating
neighborhood information during balancing: No GIN$_T$ can
remain competitive at moderate $\beta_{XT}$ values, particularly on
BC and Flickr, but HINet performs best across all datasets under the strongest covariate-dependent treatment assignment. On Coauthor-CS, No
GIN$_T$ performs worse than both HINet and the $\alpha=0$
variant for all $\beta_{XT} > 0$. These results suggest that HINet's balancing is most consistently beneficial under strong covariate-dependent treatment assignment,
and that its network-aware formulation matters most in this regime.


\textbf{Sensitivity to the balancing weight $\alpha$ and tolerance $p$.}\;
The panels in the top row of \Cref{fig:sensitivity} show that the
relation between $\alpha$ and the estimation error is non-monotonic and
dataset-dependent. Moderate values reduce mean errors by more than half on Homophily Sim and also improve performance on Flickr and BA Sim. The gains are smaller on BC, while Coauthor-CS shows no consistent
improvement over $\alpha=0$. Large values of $\alpha$ increase counterfactual error on
several datasets, reflecting the trade-off between reducing
representation discrepancy and retaining outcome-relevant information.
The panels in the bottom row show the sensitivity of estimation error to $p$, the maximum increase in factual validation loss relative to $\alpha=0$. For most datasets, performance is stable for moderate values of $p$ and less sensitive than when $\alpha$ is set directly. Higher tolerances can nevertheless increase error on some datasets. The heuristic relies on observable quantities, but its informativeness depends on whether the factual validation loss varies meaningfully with $\alpha$. \Cref{sec:failure_modes_heuristic} examines this limitation using two DGPs in which this variation is weak and the selected values of $\alpha$ increase counterfactual error.

\textbf{GNN architectures.}\;
HINet's GNN aggregator is modular: we replace GIN with GAT \citep{velickovic2017graph}, GraphSAGE \citep{hamilton2017inductive}, and the GCN architecture used by prior spillover-effect estimators \citep{jiang2022estimating,chendoubly}, evaluating each under the (default) feature-weighted mean and sum exposure mappings. \Cref{app:gnn_arch} reports the results. None of the architectures performs uniformly best. Under the feature-weighted mean mapping, GraphSAGE generally has the lowest reported mean error, with GIN being the second-best in every comparison. Under the sum mapping, GIN has the lowest reported mean error in all but one comparison, 
with GCN's errors now closer to GIN's,
while GAT and GraphSAGE perform considerably worse. This is consistent with GIN's theoretical property of preserving information relevant to count-based exposure mappings \citep{xu2018powerful}. Because the true mechanism is generally unknown in practice, GIN's consistent performance across mappings makes it the preferred default choice.
\section{Conclusion}
We introduced HINet, a method for estimating heterogeneous treatment
effects in networks that does not require a prespecified exposure mapping.
HINet uses an expressive GNN to model how neighborhood
treatments and covariates affect outcomes and relies on network-aware adversarial
learning to balance the learned representations. 
We derived a generalization bound over local treatment configurations and introduced the evaluation metrics CNEE and PEHNE. 
HINet performs consistently well across exposure mappings and datasets considered,
whereas methods that rely on a prespecified exposure
mapping can deteriorate sharply when it is misspecified. Because the mechanisms driving interference are typically difficult to specify in real-world applications, this consistency makes HINet a strong choice in practice.

\bibliographystyle{iclr2026_conference}
\bibliography{references}

\clearpage
\appendix
\crefalias{section}{appendix}
\crefalias{subsection}{appendix}

\section{Causal Structure under Homophily}\label{app:homophily_dag}

In the DAG in \Cref{fig:causal graph}, the network is treated as fixed and exogenous. Under homophily, edge formation instead depends on node covariates. Let $E_{ij}$ be the indicator that nodes $i$ and $j$ are connected. Then $\mathbf{X}_i\rightarrow E_{ij}\leftarrow\mathbf{X}_j$, and the interference arrows between $i$ and $j$ are present only when $E_{ij}=1$. The two contexts
are shown in \Cref{fig:homophily_causal_graph_side_by_side}. This is a labeled, or context-specific, DAG, meaning that the structure of the DAG depends on the value of a conditioned variable \citep{pensar2015labeled}. 
Importantly, conditioning on a node $i$'s one-hop neighborhood implicitly conditions on the edge variables in $\{E_{ik}\}_{k\in \mathcal{V}}$, which act as colliders, inducing an association between the features $\mathbf{X}_i$ and $\mathbf{X}_k$ of these nodes \citep{pearl2009causality}.
Under strong ignorability, the induced association does not
invalidate identification: the potential outcomes remain independent of the
local treatment configuration conditional on
$\mathbf{C}_i=(\mathbf{X}_i,\mathbf{X}_{\mathcal{N}_i})$.
\begin{figure}[!htbp]
    \centering
    \begin{subfigure}[b]{0.30\textwidth}
        \centering
        \resizebox{\linewidth}{!}{\begin{tikzpicture}[>=stealth, node distance=1.5cm, on grid, auto,
        Xnode/.style={circle,draw,fill = cyan!20,minimum size=1.5cm,font=\LARGE},
        Tnode/.style={circle,draw, fill = green!20,minimum size=1.5cm,font=\LARGE},
        Ynode/.style={circle,draw, fill = yellow!20,minimum size=1.5cm,font=\LARGE},
        Enode/.style={circle,draw, fill = orange!20,minimum size=1.5cm,font=\LARGE},
every node/.style={circle, draw, minimum size=1.5cm, inner sep=0pt}
        every path/.style={thick}]  
        
        \node[Xnode] (Xi) {$\mathbf{X}_i$};
        \node[Xnode, below=4cm of Xi] (Xj) {$\mathbf{X}_j$};

        \node[Enode, below right=2cm and 1cm of Xi](Eij) {${E}_{ij}$};
        \node[Tnode, right=3cm of Xi] (Ti) {$T_i$};
        \node[Tnode, below=4cm of Ti] (Tj) {$T_j$};

        \node[Ynode, right=3cm of Ti] (Yi) {$Y_i$};
        \node[Ynode, below=4cm of Yi] (Yj) {$Y_j$};
        
            \path[->] 
                    (Xi) edge[draw=orange!50] (Eij)
                    (Xj) edge[draw=orange!50] (Eij)
                    (Xi) edge (Ti)
                     (Xi) edge[draw=orange!50,bend left=15] (Tj)
                     (Xj) edge[bend right=15,draw=orange!50] (Ti)
                     
                     (Xj) edge (Tj)  
                     (Xi) edge[bend left] (Yi)
                     (Xi) edge[draw=orange!50,opacity=1] (Yj)  
                     (Xj) edge[draw=orange!50] (Yi)
                     (Xj) edge[bend right, opacity=1] (Yj)  
                    
                     (Ti) edge (Yi)
                     (Ti) edge[draw=orange!50,opacity=1] (Yj)  
                     (Tj) edge[draw=orange!50,] (Yi)  
                     (Tj) edge[opacity=1] (Yj);  
\end{tikzpicture}}
        \caption{$E_{ij} = 1$}
        \label{fig:homophily_causal_graph_a}
    \end{subfigure}
    \hspace{0.05\textwidth}
    \begin{subfigure}[b]{0.30\textwidth}
        \centering
        \resizebox{\linewidth}{!}{\begin{tikzpicture}[>=stealth, node distance=1.5cm, on grid, auto,
        Xnode/.style={circle,draw,fill = cyan!20,minimum size=1.5cm,font=\LARGE},
        Tnode/.style={circle,draw, fill = green!20,minimum size=1.5cm,font=\LARGE},
        Ynode/.style={circle,draw, fill = yellow!20,minimum size=1.5cm,font=\LARGE},
        Enode/.style={circle,draw, fill = orange!20,minimum size=1.5cm,font=\LARGE},
every node/.style={circle, draw, minimum size=1.5cm, inner sep=0pt}
        every path/.style={very thick}]  
        
        \node[Xnode] (Xi) {$\mathbf{X}_i$};
        \node[Xnode, below=4cm of Xi] (Xj) {$\mathbf{X}_j$};

        \node[Enode, below right=2cm and 1cm of Xi](Eij) {${E}_{ij}$};
        \node[Tnode, right=3cm of Xi] (Ti) {$T_i$};
        \node[Tnode, below=4cm of Ti] (Tj) {$T_j$};

        \node[Ynode, right=3cm of Ti] (Yi) {$Y_i$};
        \node[Ynode, below=4cm of Yi] (Yj) {$Y_j$};
        
            \path[->] 
                    (Xi) edge[draw=orange!50] (Eij)
                    (Xj) edge[draw=orange!50] (Eij)
                    (Xi) edge (Ti)
                     
                     (Xj) edge (Tj)  
                     (Xi) edge[bend left] (Yi)
                     (Xj) edge[bend right, opacity=1] (Yj)  
                    
                     (Ti) edge (Yi)
                     (Tj) edge[opacity=1] (Yj);  
\end{tikzpicture}}
        \caption{$E_{ij} = 0$}
        \label{fig:homophily_causal_graph_b}
    \end{subfigure}
    \caption{DAGs representing the causal structure when homophily is present. $E_{ij}$ is the binary variable which is 1 if there is an edge between node $i$ and $j$. Conditioning on the presence of an edge ($E_{ij}$) reveals the underlying causal structure.}
    \label{fig:homophily_causal_graph_side_by_side}
\end{figure}
\clearpage
\section{Proof of \texorpdfstring{\Cref{theorem:neighbors}}{Proposition 1}}\label{app:proof_prop1}

\begin{restateprop}{theorem:neighbors}
 Assume that
\[
p(\boldsymbol{\phi}_k,\boldsymbol{\phi}_{\mathcal{N}_k}\mid T_k)
=
p(\boldsymbol{\phi}_k,\boldsymbol{\phi}_{\mathcal{N}_k})
\qquad
\forall k\in\mathcal{V}.
\]
Under the conditional independence conditions implied by the causal structure shown in \Cref{fig:causal graph}, the representations are also invariant with respect to each neighbor's treatment: \[p(\boldsymbol{\phi}_i,\boldsymbol{\phi}_{\mathcal{N}_i}\mid T_j)=p(\boldsymbol{\phi}_i,\boldsymbol{\phi}_{\mathcal{N}_i}), \forall i\in \mathcal{V}, \forall j \in \mathcal{N}_i.\]
\end{restateprop}
\begin{proof}

For any neighbor \(j\) of \(i\) define $\mathcal{A}_{ij}$ and $\mathcal{B}_{ij}$: 
\[
\mathcal{A}_{ij}:=\mathcal{N}_i\cap \mathcal{N}_j,\qquad \mathcal{B}_{ij}:=\mathcal{N}_i\setminus\big(\mathcal{A}_{ij}\cup\{j\}\big),
\]
so that \(\mathcal{N}_i = \mathcal{A}_{ij}\cup \mathcal{B}_{ij}\cup\{j\}\).

We explicitly train our model to balance the representations for each node $i$ as follows: 
\begin{equation}
\label{eq: app balancing def}
(\boldsymbol{\phi}_i,\boldsymbol{\phi}_{\mathcal{N}_i})\ \perp\ T_i,
\; \text{i.e.,}\;
p(\boldsymbol{\phi}_i,\boldsymbol{\phi}_{\mathcal{N}_i}\mid T_i)=p(\boldsymbol{\phi}_i,\boldsymbol{\phi}_{\mathcal{N}_i}) \quad \forall i \in \mathcal{V}.
\end{equation}

Additionally, from the assumed causal graph and Markov assumption, i.e., only one-hop neighbors causally affect each other, we have that the features of the nodes in set $\mathcal{B}_{ij}$---which are neighbors of node $i$ that are not
directly connected to node $j$---are independent of $T_j$:
\[
\mathbf{X}_{\mathcal{B}_{ij}} \;\perp\; T_j. 
\]
In the assumed causal graph (\Cref{fig:causal graph}) the node features
$\{\mathbf{X}_k\}_{k\in\mathcal{V}}$ are mutually independent,
and the edges do not depend on the features, so conditioning on $\mathcal{G}$
does not create an association. Because $\boldsymbol{\phi}_k = e_\phi(\mathbf{X}_k)$ is a
per-node encoder and $T_j$ depends only on
$(\mathbf{X}_j,\mathbf{X}_{\mathcal{N}_j})$, while
$\mathcal{B}_{ij}\cap(\{j\}\cup\mathcal{N}_j)=\emptyset$, the set
$\boldsymbol{\phi}_{\mathcal{B}_{ij}}$ is a function of features that are
independent of those determining $T_j$,
$\boldsymbol{\phi}_i$, $\boldsymbol{\phi}_j$, and
$\boldsymbol{\phi}_{\mathcal{A}_{ij}}$. We thus obtain the {joint}
independence:
\begin{equation}
\label{eq:markov_independence}
\boldsymbol{\phi}_{\mathcal{B}_{ij}} \;\perp\;
\bigl(T_j,\boldsymbol{\phi}_i,\boldsymbol{\phi}_j,\boldsymbol{\phi}_{\mathcal{A}_{ij}}\bigr).
\end{equation}

Write the collection $(\boldsymbol{\phi}_i,\boldsymbol{\phi}_{\mathcal{N}_i})$ as 
\[
(\boldsymbol{\phi}_i,\boldsymbol{\phi}_{\mathcal{N}_i}) \equiv (\boldsymbol{\phi}_i,\boldsymbol{\phi}_j,\boldsymbol{\phi}_{\mathcal{A}_{ij}},\boldsymbol{\phi}_{\mathcal{B}_{ij}}).
\]
Apply the chain rule to the 
joint distribution conditional on
\(T_j\):
\begin{equation}\label{eq:chain_cond}
p(\boldsymbol{\phi}_i,\boldsymbol{\phi}_j,\boldsymbol{\phi}_{\mathcal{A}_{ij}},\boldsymbol{\phi}_{\mathcal{B}_{ij}} \mid T_j)
= p(\boldsymbol{\phi}_{\mathcal{B}_{ij}}\mid \boldsymbol{\phi}_i,\boldsymbol{\phi}_j,\boldsymbol{\phi}_{\mathcal{A}_{ij}},T_j)\; p(\boldsymbol{\phi}_i,\boldsymbol{\phi}_j,\boldsymbol{\phi}_{\mathcal{A}_{ij}} \mid T_j).
\end{equation}
Because joint independence implies conditional independence, it follows from \Cref{eq:markov_independence} that
\begin{equation}
p(\boldsymbol{\phi}_{\mathcal{B}_{ij}}\mid \boldsymbol{\phi}_i,\boldsymbol{\phi}_j,\boldsymbol{\phi}_{\mathcal{A}_{ij}},T_j)
= p(\boldsymbol{\phi}_{\mathcal{B}_{ij}}\mid \boldsymbol{\phi}_i,\boldsymbol{\phi}_j,\boldsymbol{\phi}_{\mathcal{A}_{ij}}).
\end{equation}
Next, since $(\boldsymbol{\phi}_i,\boldsymbol{\phi}_j,\boldsymbol{\phi}_{\mathcal{A}_{ij}})$ is a subset of
\((\boldsymbol{\phi}_j,\boldsymbol{\phi}_{\mathcal{N}_j})\), 
\cref{eq: app balancing def} for node $j$ implies that
\begin{equation}
p(\boldsymbol{\phi}_i,\boldsymbol{\phi}_j,\boldsymbol{\phi}_{\mathcal{A}_{ij}} \mid T_j) = p(\boldsymbol{\phi}_i,\boldsymbol{\phi}_j,\boldsymbol{\phi}_{\mathcal{A}_{ij}}).
\end{equation}
Substituting these two equalities into \cref{eq:chain_cond} yields
\begin{equation}
p(\boldsymbol{\phi}_i,\boldsymbol{\phi}_j,\boldsymbol{\phi}_{\mathcal{A}_{ij}},\boldsymbol{\phi}_{\mathcal{B}_{ij}} \mid T_j)
= p(\boldsymbol{\phi}_{\mathcal{B}_{ij}}\mid \boldsymbol{\phi}_i,\boldsymbol{\phi}_j,\boldsymbol{\phi}_{\mathcal{A}_{ij}})\; p(\boldsymbol{\phi}_i,\boldsymbol{\phi}_j,\boldsymbol{\phi}_{\mathcal{A}_{ij}}).
\end{equation}
Now, applying the chain rule again, we have that
\begin{equation}
p(\boldsymbol{\phi}_i,\boldsymbol{\phi}_j,\boldsymbol{\phi}_{\mathcal{A}_{ij}},\boldsymbol{\phi}_{\mathcal{B}_{ij}} \mid T_j)
= p(\boldsymbol{\phi}_i,\boldsymbol{\phi}_j,\boldsymbol{\phi}_{\mathcal{A}_{ij}},\boldsymbol{\phi}_{\mathcal{B}_{ij}}).
\end{equation}
Therefore,
\begin{equation}
p(\boldsymbol{\phi}_i,\boldsymbol{\phi}_{\mathcal{N}_i} \mid T_j) \;=\; p(\boldsymbol{\phi}_i,\boldsymbol{\phi}_{\mathcal{N}_i}),
\end{equation}
which is the claimed independence from \(T_j\).
\end{proof}

Under these idealized conditions, \Cref{theorem:neighbors} extends 
invariance of the local representations with respect to each node’s own treatment to marginal invariance with respect to the treatments of its
neighbors.
Because the loss function trades predictive accuracy against treatment invariance, the learned representations are not guaranteed to be treatment-invariant.
This result therefore motivates the architecture rather than describing an exact property of the fitted model.

If there is homophily, the causal structure slightly changes (see \Cref{fig:homophily_causal_graph_side_by_side}). In the next section, we explain that an additional assumption on the encoder $e_\phi$ is needed for the proposition to hold. 

\clearpage
\section{Balanced Representations under Homophily}
\label{app: homophily interference}
\label{app: homophily_balancing}
\Cref{theorem:neighbors} still holds when homophily is present, but it requires an additional assumption on the encoder $e_\phi$. Due to homophily, the independence $\mathbf{X}_{\mathcal{B}_{ij}} \;\perp\; T_j$ no longer holds after we condition on one-hop neighborhoods. Conditioning on the neighborhoods of $i$ and $j$, which we do in our proof, amounts to conditioning on all edges in $\{E_{ik}\}_{k \in \mathcal{V}} \cup \{E_{jk}\}_{k \in \mathcal{V}}$. More specifically, the problem is that we condition on the binary edge indicator variables $E_{bi}$ and $E_{bj}$ for every $b \in \mathcal{B}_{ij}$. These variables act as colliders that create a pathway from each $\mathbf{X}_{b}$ to $\mathbf{X}_i$ and $\mathbf{X}_j$. Concretely, conditioning on the neighborhoods of $i$ and $j$ opens the paths
\begin{equation}
\begin{aligned}
\mathbf{X}_{b} &\rightarrow E_{bi} \leftarrow \mathbf{X}_i \rightarrow T_j \\
\text{and} \qquad \mathbf{X}_{b} &\rightarrow E_{bj} \leftarrow \mathbf{X}_j \rightarrow T_j,
\qquad \forall b \in \mathcal{B}_{ij},
\end{aligned}
\end{equation}
which creates an association between $\mathbf{X}_{\mathcal{B}_{ij}}$ and $T_j$. Note that we do not condition on other edges than the ones in the set $\{E_{ik}\}_{k \in \mathcal{V}} \cup \{E_{jk}\}_{k \in \mathcal{V}}$, meaning no associations between $b$ and other nodes in the graph are induced. Additionally, given that by construction $E_{bj}=0$ for all $b \in \mathcal{B}_{ij}$, there is no direct causal effect from  $\mathbf{X}_{\mathcal{B}_{ij}}$ to $T_j$.

Now, to recover the independence of $\mathbf{X}_{\mathcal{B}_{ij}}$ and  $T_j$ needed for the proof, we must close these paths by conditioning on $\mathbf{X}_i$  and $\mathbf{X}_j$. This yields
\[
\mathbf{X}_{\mathcal{B}_{ij}} \perp\!\!\!\perp T_j \mid \mathbf{X}_i,\mathbf{X}_{j}.
\]
This independence is stated in terms of the covariates
$\mathbf{X}_i,\mathbf{X}_j$, whereas the proof conditions on the learned
representations $\boldsymbol{\phi}_i,\boldsymbol{\phi}_j$. Because the encoder
$e_\phi$ need not be invertible, $\boldsymbol{\phi}_i$ and $\boldsymbol{\phi}_j$ may
discard the covariate information that closes these collider paths, so the
conditional independence does not transfer to the representations
automatically. We therefore make a representation-sufficiency assumption for
the treatment assignment mechanism: the representations retain the information
needed to block these paths, i.e.,
\begin{equation}
\label{eq:assignment_sufficiency}
\boldsymbol{\phi}_{\mathcal{B}_{ij}} \perp\!\!\!\perp T_j \mid
\boldsymbol{\phi}_i,\boldsymbol{\phi}_{j},\boldsymbol{\phi}_{\mathcal{A}_{ij}}.
\end{equation}
This condition is more plausible when the covariates driving edge formation are
independent of those driving treatment assignment. When the same covariates
drive both mechanisms, this condition requires the encoder to retain the relevant
assignment information, and it may hold only approximately. It is not guaranteed by
adversarial balancing, and it becomes less likely to hold when the encoder discards a
large amount of information or when homophily is latent. Throughout this paper,
we assume that there is no latent homophily.
Under~\cref{eq:assignment_sufficiency},
\[
p(\boldsymbol{\phi}_{\mathcal{B}_{ij}}\mid
\boldsymbol{\phi}_i,\boldsymbol{\phi}_j,\boldsymbol{\phi}_{\mathcal{A}_{ij}},T_j)
= p(\boldsymbol{\phi}_{\mathcal{B}_{ij}}\mid
\boldsymbol{\phi}_i,\boldsymbol{\phi}_j,\boldsymbol{\phi}_{\mathcal{A}_{ij}}).
\]
The rest of the proof proceeds identically from this point.

\clearpage
\section{A Generalization Bound for Counterfactual-Outcome and ITTE Estimation under Interference}
\label{app:bound}
We derive upper bounds on the expected counterfactual-outcome error (CNEE) and treatment-effect error (PEHNE) under one-hop interference. The argument adapts the representation-learning bounds of \citet{johansson2022generalization}, building on \citet{shalit2017estimating}, to the interference setting. The bounds involve the factual outcome risk and a discrepancy over representations and local treatment configurations. HINet targets a tractable component of this discrepancy, as discussed below.


\subsection{Setup and notation}
We use the notation of \Cref{sec:Problem setup}. Node $i$ has covariates $\mathbf{X}_i$, treatment
$T_i$, outcome $Y_i$, and one-hop neighbors $\mathcal{N}_i$. $\mathbf{X}_{\mathcal{N}_i}$
and $\mathbf{T}_{\mathcal{N}_i}$ denote the set of covariates and treatments of those neighbors.
$\boldsymbol{\phi}_k=e_\phi(\mathbf{X}_k)$ is the learned node representation for a node $k$, and
$\mathcal{M}(\mathbf{X}_i,t_i,\mathbf{X}_{\mathcal{N}_i},\mathbf{t}_{\mathcal{N}_i})$ is the
model predicting the potential outcome $Y_i(t_i,\mathbf{t}_{\mathcal{N}_i})$. To keep the
expressions readable, we group the covariates into the \emph{local covariate environment}
$\mathbf{C}_i:=(\mathbf{X}_i,\mathbf{X}_{\mathcal{N}_i})$, the treatments into the
\emph{local treatment configuration}
$\mathbf{A}_i:=(T_i,\mathbf{T}_{\mathcal{N}_i})$, and the node representations into $\boldsymbol{\Phi}_i:=(\boldsymbol{\phi}_i,\boldsymbol{\phi}_{\mathcal{N}_i})$. We write
$\mathcal{M}(\mathbf{C}_i,\mathbf{a})$ for
$\mathcal{M}(\mathbf{X}_i,t_i,\mathbf{X}_{\mathcal{N}_i},\mathbf{t}_{\mathcal{N}_i})$,
$Y_i(\mathbf{a})$ for $Y_i(t_i,\mathbf{t}_{\mathcal{N}_i})$, and
$\bar{Y}_i(\mathbf{a}):=\mathbb{E}[Y_i(\mathbf{a})\mid\mathbf{C}_i]$ for the conditional expected
potential outcome.

With the squared loss, define the pointwise loss, the marginal risk at a specific configuration
$\mathbf{a}$, and the (observable) factual risk over the observed treatment configuration distribution:
\begin{align*}
\ell_{\mathcal{M}}(\mathbf{a};\mathbf{C}_i)&:=\mathbb{E}\big[(Y_i(\mathbf{a})-\mathcal{M}(\mathbf{C}_i,\mathbf{a}))^2\mid\mathbf{C}_i\big],\\
R_{\mathbf{a}}(\mathcal{M})&:=\mathbb{E}_{\mathbf{C}_i}\big[\ell_{\mathcal{M}}(\mathbf{a};\mathbf{C}_i)\big],\\
R^{F}(\mathcal{M})&:=\mathbb{E}_{(\mathbf{C}_i,\mathbf{A}_i)}\big[\ell_{\mathcal{M}}(\mathbf{A}_i;\mathbf{C}_i)\big],
\end{align*}
where $R^{F}$ averages over the observed treatment configurations $\mathbf{A}_i\sim p(\mathbf{A}_i\mid\mathbf{C}_i)$. Let $\sigma_{\mathbf{a}}^2:=\mathbb{E}_{\mathbf{C}_i}[\mathrm{Var}(Y_i(\mathbf{a})\mid\mathbf{C}_i)]$, and the 
Individual Total Treatment Effect (ITTE) and its estimate are
$\omega_i(\mathbf{a})=\bar{Y}_i(\mathbf{a})-\bar{Y}_i(\mathbf{0})$ and
$\hat{\omega}_i(\mathbf{a})=\mathcal{M}(\mathbf{C}_i,\mathbf{a})-\mathcal{M}(\mathbf{C}_i,\mathbf{0})$ \citep{caljon2025optimizing}.
PEHNE and CNEE are \emph{population} losses: the expected squared ITTE and
counterfactual-outcome errors over the local covariate environments and over a target distribution $\rho$ of
counterfactual treatment configurations,
\begin{align*}
\epsilon_{\mathrm{CNEE}}(\mathcal{M})&:=\mathbb{E}_{\mathbf{C}_i}\mathbb{E}_{\mathbf{a}\sim\rho}\big[(\bar{Y}_i(\mathbf{a})-\mathcal{M}(\mathbf{C}_i,\mathbf{a}))^2\big],\\
\epsilon_{\mathrm{PEHNE}}(\mathcal{M})&:=\mathbb{E}_{\mathbf{C}_i}\mathbb{E}_{\mathbf{a}\sim\rho}\big[(\omega_i(\mathbf{a})-\hat{\omega}_i(\mathbf{a}))^2\big].
\end{align*}
Here $\rho$ specifies how counterfactual networks are sampled. Because nodes have different neighborhoods, this sampling scheme induces a different distribution of local treatment configurations for each node. We use $\mathbf{a}\sim\rho$ to denote a local treatment configuration generated by this scheme. The $m$ counterfactual networks sampled in the experiments approximate the expectation with respect to $\rho$.
Finally, let
$p^{\mathrm{obs}}_{\boldsymbol{\Phi}}$ denote the distribution of $(\boldsymbol{\Phi}_i,\mathbf{A}_i)$
under the observed treatment assignment, $p^{\rho}_{\boldsymbol{\Phi}}$ the distribution of
$(\boldsymbol{\Phi}_i,\mathbf{a})$ when $\mathbf{a}\sim\rho$ is drawn independently of
$\mathbf{C}_i$, and $p^{\mathbf{0}}_{\boldsymbol{\Phi}}$ the distribution of
$(\boldsymbol{\Phi}_i,\mathbf{0})$
for the zero treatment configuration $\mathbf{a}=\mathbf{0}$.

\subsection{Assumptions}
We adopt the assumptions of \Cref{sec:Problem setup}, written for a configuration $\mathbf{a}$:
\textbf{(A1)} consistency;
\textbf{(A2)} overlap, $\delta<p(\mathbf{A}_i=\mathbf{a}\mid\mathbf{C}_i)<1-\delta$ on the
support of $\rho$ and at $\mathbf{0}$;
\textbf{(A3)} strong ignorability,
$Y_i(\mathbf{a})\perp\!\!\!\perp(T_i,\mathbf{T}_{\mathcal{N}_i})\mid\mathbf{C}_i$;
\textbf{(A4)} existence of an exposure mapping $z(\cdot)$ such that $Y_i(\mathbf{a})$ depends
on $\mathbf{a}$ only through $(t_i,z(\mathbf{t}_{\mathcal{N}_i},\mathbf{X}_{\mathcal{N}_i}))$
\citep{aronow2017estimating,forastiere2021identification,savje2024misspecified}, which we do
not know.
For the representation, two conditions need to be added:
\textbf{(A5)} \emph{representation sufficiency}: the conditional loss depends on
$\mathbf{C}_i$ only through $\boldsymbol{\Phi}_i$, i.e., \
$\ell_{\mathcal{M}}(\mathbf{a};\mathbf{C}_i)=\tilde{\ell}_{\mathcal{M}}(\mathbf{a};\boldsymbol{\Phi}_i)$
(implied by $Y_i(\mathbf{a})\perp\!\!\!\perp\mathbf{C}_i\mid\boldsymbol{\Phi}_i$). This assumption replaces invertibility. In  \citet{johansson2022generalization}, invertibility is used as a convenient condition for a clean change of
variables, not as a requirement: their balancing
result holds for any $\Phi$ with $\ell_f\perp \mathbf{X}\mid\Phi(\mathbf{X})$. Invertibility would also conflict with the
purpose of adversarial balancing. When covariates and treatment are dependent, an invertible encoder preserves that dependence and therefore cannot produce exact treatment invariance.
Balancing necessarily discards some information. Assumption~(A5) permits such information loss if the discarded information is irrelevant to outcome prediction. It fails when balancing removes outcome-relevant information, potentially leading to representation-induced confounding \citep{melnychuk2024bounds}. 
\textbf{(A6)} \emph{IPM regularity}: for the function class $\mathcal{L}$ defining the
discrepancy $\mathrm{IPM}_{\mathcal{L}}(P,Q)=\sup_{g\in\mathcal{L}}|\mathbb{E}_P[g]-\mathbb{E}_Q[g]|$,
there is a constant $C_{\boldsymbol{\Phi}}>0$ such that the joint map
$(\boldsymbol{\Phi},\mathbf{a})\mapsto\tilde{\ell}_{\mathcal{M}}(\mathbf{a};\boldsymbol{\Phi})/C_{\boldsymbol{\Phi}}$
belongs to $\mathcal{L}$.
This is the standard regularity condition of IPM-based
bounds
\citep{johansson2022generalization,shalit2017estimating}.

\textbf{Interpretation of overlap.}\;
Assumption (A2) is a population-level support condition: it does not require every local treatment configuration to appear in the observed data. In our data-generating processes (DGPs), the Bernoulli propensities are strictly between zero and one, so every local configuration has positive probability, although some are very unlikely. As discussed in \Cref{sec:Problem setup}, treatment-configuration imbalance arises because the distribution of $\mathbf{A}_i$ can vary with $\mathbf{C}_i$, so that some configurations are rarely observed for particular local covariate environments. The corresponding potential outcomes remain identifiable under our assumptions, but the model must extrapolate further beyond the observed data to estimate them. Since each node is observed under only one configuration, HINet addresses this by learning a shared outcome model across nodes and configurations and reducing the (potential) dependence between learned node representations and the treatment assignment using domain-adversarial training.


\subsection{Counterfactual-error (CNEE) bound}
We begin by bounding the risk under an arbitrary target distribution over treatment configurations in terms of the observed factual risk and a representation discrepancy.

\begin{lemma}
\label{lem:ipm}
Under (A1)--(A6), for any distribution $\rho$ over treatment configurations,
\begin{equation*}
\mathbb{E}_{\mathbf{a}\sim\rho}[R_{\mathbf{a}}(\mathcal{M})]\le R^{F}(\mathcal{M})+C_{\boldsymbol{\Phi}}\,\mathrm{IPM}_{\mathcal{L}}\big(p^{\rho}_{\boldsymbol{\Phi}},p^{\mathrm{obs}}_{\boldsymbol{\Phi}}\big).
\end{equation*}
\end{lemma}
\begin{proof}
Both risks,
$R_{\mathbf{a}}(\mathcal{M})$ and $R^{F}(\mathcal{M})$,
integrate the same loss for different joint distributions of
$(\mathbf{C}_i,\mathbf{a})$: the target distribution induced
by $\rho$ and the distribution under the observed treatment assignment. By (A5), the
loss depends on $\mathbf{C}_i$ only through
$\boldsymbol{\Phi}_i$. Mapping both distributions through the encoder therefore
gives
\begin{align*}
&\mathbb{E}_{\mathbf{a}\sim\rho}[R_{\mathbf{a}}(\mathcal{M})]
-
R^{F}(\mathcal{M})
\\
&\quad =
\int
\tilde{\ell}_{\mathcal{M}}(\mathbf{a};\boldsymbol{\Phi})
\left(
p^{\rho}_{\boldsymbol{\Phi}}(\boldsymbol{\Phi},\mathbf{a})
-
p^{\mathrm{obs}}_{\boldsymbol{\Phi}}(\boldsymbol{\Phi},\mathbf{a})
\right)
\,d\boldsymbol{\Phi}\,d\mathbf{a}
\\
&\quad \leq
C_{\boldsymbol{\Phi}}
\operatorname{IPM}_{\mathcal{L}}
\left(
p^{\rho}_{\boldsymbol{\Phi}},
p^{\mathrm{obs}}_{\boldsymbol{\Phi}}
\right),
\end{align*}
where the inequality follows from (A6). The supremum is $\mathrm{IPM}_{\mathcal{L}}\big(p^{\rho}_{\boldsymbol{\Phi}},p^{\mathrm{obs}}_{\boldsymbol{\Phi}}\big)$ by definition \citep{muller1997integral}.
\end{proof}

The main differences from \citet{johansson2022generalization} are (i) representation sufficiency
instead of invertibility, and (ii) a single IPM between joint
distributions over the space of (representation, treatment configuration) pairs instead of pairwise treated--control
distances.

\begin{restateprop}{prop:bound}
\emph{(CNEE bound.)} Under (A1)--(A6),
\begin{equation*}
\epsilon_{\mathrm{CNEE}}(\mathcal{M})\le R^{F}(\mathcal{M})+C_{\boldsymbol{\Phi}}\,\mathrm{IPM}_{\mathcal{L}}\big(p^{\rho}_{\boldsymbol{\Phi}},p^{\mathrm{obs}}_{\boldsymbol{\Phi}}\big)-\mathbb{E}_{\mathbf{a}\sim\rho}[\sigma_{\mathbf{a}}^2].
\end{equation*}
\end{restateprop}
\begin{proof}
The conditional bias--variance decomposition of $\ell_{\mathcal{M}}(\mathbf{a};\mathbf{C}_i)$ gives
\[\ell_{\mathcal{M}}(\mathbf{a};\mathbf{C}_i)=(\bar{Y}_i(\mathbf{a})-\mathcal{M}(\mathbf{C}_i,\mathbf{a}))^2+\mathrm{Var}(Y_i(\mathbf{a})\mid\mathbf{C}_i).\]
Now, taking the expectation with respect to $\mathbf{C}_i$ gives
\[R_{\mathbf{a}}(\mathcal{M})= \mathbb{E}_{\mathbf{C}_i}[(\bar{Y}_i(\mathbf{a})-\mathcal{M}(\mathbf{C}_i,\mathbf{a}))^2]+\sigma_{\mathbf{a}}^2,\]
or equivalently,
\[\mathbb{E}_{\mathbf{C}_i}[(\bar{Y}_i(\mathbf{a})-\mathcal{M}(\mathbf{C}_i,\mathbf{a}))^2]=R_{\mathbf{a}}(\mathcal{M})-\sigma_{\mathbf{a}}^2.\]
Taking the expectation over $\mathbf{a}\sim\rho$ gives
\[\epsilon_{\mathrm{CNEE}}(\mathcal{M})=\mathbb{E}_{\mathbf{a}\sim\rho}[R_{\mathbf{a}}(\mathcal{M})]-\mathbb{E}_{\mathbf{a}\sim\rho}[\sigma_{\mathbf{a}}^2].\]
Now, apply Lemma~\ref{lem:ipm}: 
\[\epsilon_{\mathrm{CNEE}}(\mathcal{M})\le R^{F}(\mathcal{M})+C_{\boldsymbol{\Phi}}\,\mathrm{IPM}_{\mathcal{L}}\big(p^{\rho}_{\boldsymbol{\Phi}},p^{\mathrm{obs}}_{\boldsymbol{\Phi}}\big)-\mathbb{E}_{\mathbf{a}\sim\rho}[\sigma_{\mathbf{a}}^2].\]
\end{proof}

The $-\mathbb{E}_{\mathbf{a}\sim\rho}[\sigma_{\mathbf{a}}^2]$
term is an irreducible-noise correction: $R_{\mathbf{a}}$ is the error for the noisy outcome
$Y_i(\mathbf{a})$, whereas CNEE measures error against the conditional expectation
$\bar{Y}_i(\mathbf{a})=\mathbb{E}[Y_i(\mathbf{a})\mid\mathbf{C}_i]$. 
This correction does not imply that noise aids estimation. In \Cref{prop:bound} we write $\mathbb{E}_{\mathbf{a}\sim\rho}[\sigma_{\mathbf{a}}^2]$ as $\sigma^2$ for simplicity. 

\subsection{ITTE (PEHNE) bound}
The ITTE is a contrast against the zero treatment configuration, so its error inherits a second
discrepancy term.

\begin{lemma}
\label{lem:pehe}
Under (A1)--(A4), with squared loss,
\begin{equation*}
\epsilon_{\mathrm{PEHNE}}(\mathcal{M})
\leq 2\,\mathbb{E}_{\mathbf{a}\sim\rho}\bigl[R_{\mathbf{a}}(\mathcal{M})\bigr]
+ 2\,R_{\mathbf{0}}(\mathcal{M})
- 2\left(\mathbb{E}_{\mathbf{a}\sim\rho}\bigl[\sigma_{\mathbf{a}}^2\bigr] + \sigma_{\mathbf{0}}^2\right).
\end{equation*}
\end{lemma}

\begin{proof}
Since $\omega_i(\mathbf{a})-\hat{\omega}_i(\mathbf{a})=(\bar{Y}_i(\mathbf{a})-\mathcal{M}(\mathbf{C}_i,\mathbf{a}))-(\bar{Y}_i(\mathbf{0})-\mathcal{M}(\mathbf{C}_i,\mathbf{0}))$,
the relaxed triangle inequality $(u-v)^2\le 2u^2+2v^2$ gives
\begin{equation*}
\bigl(\omega_i(\mathbf{a})-\hat{\omega}_i(\mathbf{a})\bigr)^2
\leq
2\bigl(\bar{Y}_i(\mathbf{a})-\mathcal{M}(\mathbf{C}_i,\mathbf{a})\bigr)^2
+2\bigl(\bar{Y}_i(\mathbf{0})-\mathcal{M}(\mathbf{C}_i,\mathbf{0})\bigr)^2.
\end{equation*}
Taking the expectations $\mathbb{E}_{\mathbf{C}_i}$ and $\mathbb{E}_{\mathbf{a}\sim\rho}$, and using
\[
\mathbb{E}_{\mathbf{C}_i}[(\bar{Y}_i(\mathbf{a})-\mathcal{M}(\mathbf{C}_i,\mathbf{a}))^2]=R_{\mathbf{a}}(\mathcal{M})-\sigma_{\mathbf{a}}^2,\]
\[\mathbb{E}_{\mathbf{C}_i}\!\left[(\bar Y_i(0)-\mathcal{M}(\mathbf{C}_i,0))^2\right]
    = R_0(\mathcal{M})-\sigma_0^2,
\]
gives the stated result.
\end{proof}
The factor 2 comes from the relaxed triangle inequality, while the two risk terms correspond to the target treatment configuration and the zero treatment configuration.

\begin{proposition}[ITTE bound under interference]
\label{thm:bound}
Under (A1)--(A6), with $\bar{\sigma}^2:=\mathbb{E}_{\mathbf{a}\sim\rho}[\sigma_{\mathbf{a}}^2]+\sigma_{\mathbf{0}}^2$,
\begin{equation}
\label{eq:bound}
\epsilon_{\mathrm{PEHNE}}(\mathcal{M})\le 4R^{F}(\mathcal{M})-2\bar{\sigma}^2
+2C_{\boldsymbol{\Phi}}\Big[\mathrm{IPM}_{\mathcal{L}}\big(p^{\rho}_{\boldsymbol{\Phi}},p^{\mathrm{obs}}_{\boldsymbol{\Phi}}\big)
+\mathrm{IPM}_{\mathcal{L}}\big(p^{\mathbf{0}}_{\boldsymbol{\Phi}},p^{\mathrm{obs}}_{\boldsymbol{\Phi}}\big)\Big].
\end{equation}
\end{proposition}

\begin{proof}

First, applying Lemma~\ref{lem:ipm} to the target distribution $\rho$
gives
\begin{equation}
\mathbb{E}_{\mathbf{a}\sim\rho}
    [R_{\mathbf{a}}(\mathcal{M})]
\leq
R^{F}(\mathcal{M})
+
C_{\boldsymbol{\Phi}}\,
\mathrm{IPM}_{\mathcal{L}}
\left(
p^{\rho}_{\boldsymbol{\Phi}},
p^{\mathrm{obs}}_{\boldsymbol{\Phi}}
\right).
\label{eq:rho-risk-bound}
\end{equation}

Second, let $\delta_{\mathbf{0}}$ denote the point mass at the zero treatment configuration, so that
$\mathbf{a}=\mathbf{0}$ almost surely when
$\mathbf{a}\sim\delta_{\mathbf{0}}$. Hence,
\[
\mathbb{E}_{\mathbf{a}\sim\delta_{\mathbf{0}}}
    [R_{\mathbf{a}}(\mathcal{M})]
=
R_{\mathbf{0}}(\mathcal{M}).
\]
Since Lemma~\ref{lem:ipm} holds for any distribution $\rho$ over
treatment configurations, we can apply it to
$\rho = \delta_{\mathbf{0}}$.
This gives
\begin{equation}
R_{\mathbf{0}}(\mathcal{M})
\leq
R^{F}(\mathcal{M})
+
C_{\boldsymbol{\Phi}}\,
\mathrm{IPM}_{\mathcal{L}}
\left(
p^{\mathbf{0}}_{\boldsymbol{\Phi}},
p^{\mathrm{obs}}_{\boldsymbol{\Phi}}
\right).
\label{eq:zero-risk-bound}
\end{equation}

Substituting \eqref{eq:rho-risk-bound} and
\eqref{eq:zero-risk-bound} into
\Cref{lem:pehe} gives the stated result.
\end{proof}

\subsection{Consequences}

\paragraph{The CNEE bound is the cleanest alignment with
HINet's objective.}
\Cref{prop:bound} has the same structure as HINet's
objective: the outcome loss $\mathcal{L}_y$ estimates the
factual risk $R^{F}(\mathcal{M})$, while the
gradient-reversal term $\alpha\mathcal{L}_t$ targets a
tractable proxy for the representation discrepancy. For the
ITTE, the PEHNE bound \eqref{eq:bound} additionally contains
the 
discrepancy
$\mathrm{IPM}_{\mathcal{L}}
\left(
p^{\mathbf{0}}_{\boldsymbol{\Phi}},
p^{\mathrm{obs}}_{\boldsymbol{\Phi}}
\right)$.

\paragraph{The adversarial objective as a proxy for treatment-configuration-level balancing.}
The IPM compares joint distributions of the representation and treatment
configuration. It may therefore reflect both dependence between
$\boldsymbol{\Phi}_i$ and $\mathbf{A}_i$, and differences in how frequently
treatment configurations occur under the observed and target assignments. Representation
balancing targets the former but does not alter the frequencies of
treatment configurations.

HINet does not directly minimize the treatment-configuration-level discrepancy in
\Cref{prop:bound}. In principle, the discriminator could target the local
treatment configuration
$\mathbf{A}_i$
instead of a node's own treatment $T_i$. For a node with degree
$|\mathcal{N}_i|$, however, $\mathbf{A}_i$ can take up to
$2^{|\mathcal{N}_i|+1}$ values. Each node is observed under only one of these
treatment configurations, and neighborhood sizes vary across nodes. Reweighting faces
the same problem because the joint propensity $p(\mathbf{A}_i\mid\mathbf{C}_i)$ is defined over the same exponentially large space of local treatment configurations.

Existing treatment-configuration-level methods make this problem tractable by assuming
an exposure mapping that summarizes $\mathbf{A}_i$ in a lower-dimensional
quantity, such as the proportion of treated neighbors
\citep{forastiere2021identification,chendoubly}. Balancing with respect to
this quantity requires the exposure mapping to be specified in advance.
HINet instead balances with respect to each node's own treatment. This yields
a binary treatment prediction task and does not require
a prespecified exposure mapping.

In the idealized case of exact marginal invariance, applying this objective
at node $i$ yields
\[
\boldsymbol{\Phi}_i\perp T_i.
\]
Since the same objective is applied at every node, the conditions of
\Cref{theorem:neighbors} further imply
\[
\boldsymbol{\Phi}_i\perp T_j,
\qquad
\forall j\in\mathcal{N}_i.
\]
The adversarial objective therefore targets the dependence between the
representation and each individual component of the local treatment
configuration. Under exact marginal invariance, the representation is independent of each treatment indicator separately but may still depend on their joint configuration.

Marginal balancing is naturally aligned with additive exposure
mappings of the form
\[
g_i(\mathbf{T}_{\mathcal{N}_i})
=
\sum_{k\in\mathcal{N}_i}w_{ik}T_k,
\]
because the exposure is constructed from the individual treatment indicators
targeted by the adversarial objective. For exposure mappings containing
explicit interactions, such as $T_jT_k$, marginal balancing does not directly
target dependence that is present only in particular treatment combinations.
This distinction does not make marginal balancing sufficient for additive
exposure mappings. In general,
\[
\boldsymbol{\Phi}_i\perp T_k,
\qquad
\forall k\in\mathcal{N}_i,
\]
does not imply
\[
\boldsymbol{\Phi}_i\perp
g_i(\mathbf{T}_{\mathcal{N}_i}),
\]
since the representation may remain associated with the joint treatment
pattern.

Our adversarial objective is therefore a tractable proxy for the
treatment-configuration-level discrepancy in \Cref{prop:bound}, rather than a direct
minimization of that discrepancy. It targets marginal
representation--treatment dependence without requiring estimation of the 
joint propensity or prior specification of an exposure mapping. Dependence
carried only by joint treatment patterns may remain. Designing a tractable
objective that directly targets the dependence between
$\boldsymbol{\Phi}_i$ and $\mathbf{A}_i$ is therefore an interesting direction for
future work.

\paragraph{The balancing weight $\alpha$.}
Increasing $\alpha$ places greater emphasis on marginal balancing but may also lead to discarding information needed for outcome prediction, increasing $R^{F}(\mathcal{M})$, and potentially violating representation sufficiency \citep{melnychuk2024bounds}. A moderate value is therefore plausible, but is not implied by \Cref{prop:bound}. This is examined in the sensitivity analysis in \Cref{fig:sensitivity}.



\paragraph{Relation to \citet{cai2023generalization}.}  Closest to our result, \citet{cai2023generalization} derive a generalization bound for treatment effect estimation in networks: a PEHE-type error is bounded by a propensity-reweighted factual risk plus an IPM term capturing the residual dependence between the (reweighted) representations and the treatment-exposure pair $(t_i,z_i)$. Structurally, their bound is analogous to ours: a factual risk term, an IPM term, and an irreducible-noise term. There are, however, two substantial differences. First, their analysis is developed for the setting in which the exposure mapping is known: following \citet{forastiere2021identification}, potential outcomes, the joint propensity score, and the bound itself are formulated in terms of a summary of the neighbors' treatments $z_i$ (the proportion of treated neighbors). Our bound is instead stated at the level of treatment configurations $\mathbf{A}_i$
so that it remains applicable when the exposure mapping is unknown. Second, their factual and counterfactual errors average treatment-exposure pairs over the observed marginal $p(t,z)$, a natural choice for their estimands. In contrast, $\epsilon_{\mathrm{CNEE}}$ and $\epsilon_{\mathrm{PEHNE}}$ are defined with respect to a target distribution $\rho$ over counterfactual networks. 

\clearpage
\section{Performance Metrics}\label{app: Performance metrics}

\Cref{sec:Problem setup} introduces CNEE and PEHNE as evaluation metrics for counterfactual-outcome and ITTE estimation, respectively. In this section, we distinguish the population metrics defined in \Cref{app:bound} from the finite-sample approximations used in the experiments. Both metrics are inspired by the Mean Integrated Squared Error (MISE) \citep{schwab2020learning}.

\paragraph{Population metrics.}
\Cref{app:bound} defines CNEE and PEHNE as population losses under a target distribution $\rho$, which specifies how counterfactual networks are sampled. Each draw from $\rho$ yields a treatment configuration $(t_i,\mathbf{t}_{\mathcal{N}_i})$ for every node. We write $\mathbf{a}\sim\rho$ for the resulting treatment configuration.

Recall from \Cref{app:bound} that
$\bar{Y}_i(\mathbf{a})
:=
\mathbb{E}\!\left[
Y_i(\mathbf{a})
\mid
\mathbf{C}_i
\right]$,
with $\mathbf{C}_i:=(\mathbf{X}_i,\mathbf{X}_{\mathcal{N}_i})$,
so that the ITTE and its estimate are
$\omega_i(\mathbf{a})
=
\bar{Y}_i(\mathbf{a})
-
\bar{Y}_i(\mathbf{0})$
and 
$\hat{\omega}_i(\mathbf{a})
=
\mathcal{M}(\mathbf{C}_i,\mathbf{a})
-
\mathcal{M}(\mathbf{C}_i,\mathbf{0})$.

The population metrics can then be defined as
\begin{align}
\epsilon_{\mathrm{CNEE}}(\mathcal{M})
&:=
\mathbb{E}_{\mathbf{C}_i}
\mathbb{E}_{\mathbf{a}\sim\rho}
\left[
\left(
\bar{Y}_i(\mathbf{a})
-
\mathcal{M}(\mathbf{C}_i,\mathbf{a})
\right)^2
\right],
\\
\epsilon_{\mathrm{PEHNE}}(\mathcal{M})
&:=
\mathbb{E}_{\mathbf{C}_i}
\mathbb{E}_{\mathbf{a}\sim\rho}
\left[
\left(
\omega_i(\mathbf{a})
-
\hat{\omega}_i(\mathbf{a})
\right)^2
\right].
\end{align}
CNEE therefore measures expected counterfactual-outcome estimation error, whereas PEHNE measures expected ITTE estimation error. Both are population quantities that average over local covariate environments $\mathbf{C}_i$ and counterfactual networks sampled according to $\rho$.

\paragraph{Finite-sample approximations.}
In the experiments, we approximate the expectation under $\rho$ using $m$ sampled counterfactual networks and the expectation over $\mathbf{C}_i$ by averaging over the test nodes in each counterfactual network.

We instantiate $\rho$ using treatment rates that span the interval from zero to one. For $j=1,\ldots,m$, we set
\[
p_j
=
\frac{j-1}{m-1},
\qquad
n_j
=
\left\lfloor
p_j|\mathcal{V}|
\right\rfloor,
\]
and sample a set $\mathcal{S}_j\subseteq\mathcal{V}$ of $n_j$ treated nodes uniformly without replacement. We then set
\[
t_i^j
=
\mathbb{I}(i\in\mathcal{S}_j)
\qquad
\forall i\in\mathcal{V},
\]
and denote the resulting local treatment configuration of node $i$ by
\[
\mathbf{a}_i^j
:=
\left(
t_i^j,
\mathbf{t}_{\mathcal{N}_i}^j
\right).
\]
This procedure gives equal weight to the $m$ treatment rates. Conditional on a treatment rate, the treated nodes are sampled independently of their covariates and position in the network.

Let $\bar{y}_i(\mathbf{a})$ denote the ground-truth conditional mean for test node $i$ under configuration $\mathbf{a}$, and let
\[
\hat{y}_i(\mathbf{a})
:=
\mathcal{M}(\mathbf{c}_i,\mathbf{a})
\]
denote the corresponding model prediction. The finite-sample approximations are then 
\begin{align}
\widehat{\epsilon}_{\mathrm{CNEE}}^{(m)}(\mathcal{M})
&=
\frac{1}{m|\mathcal{V}|}
\sum_{j=1}^{m}
\sum_{i\in\mathcal{V}}
\left(
\bar{y}_i(\mathbf{a}_i^j)
-
\hat{y}_i(\mathbf{a}_i^j)
\right)^2,
\\
\widehat{\epsilon}_{\mathrm{PEHNE}}^{(m)}(\mathcal{M})
&=
\frac{1}{m|\mathcal{V}|}
\sum_{j=1}^{m}
\sum_{i\in\mathcal{V}}
\left(
\omega_i(\mathbf{a}_i^j)
-
\hat{\omega}_i(\mathbf{a}_i^j)
\right)^2.
\end{align}
The average over the sampled counterfactual networks is a Monte Carlo approximation of the expectation under $\rho$. 

For clarity, \Cref{alg: CNEE,alg: PEHNE} present an equivalent calculation by first computing the mean squared error (MSE) within each sampled counterfactual network and then averaging over the $m$ networks. For readability, we refer to these finite-sample approximations as CNEE and PEHNE throughout the experimental results.

\begin{algorithm}
\caption{CNEE calculation}
\label{alg: CNEE}
\begin{algorithmic}[1]
    \For{$j=1,\ldots,m$}
        \State Set
        $p_j=\frac{j-1}{m-1}$ and
        $n_j=\left\lfloor p_j|\mathcal{V}|\right\rfloor$
        \State Sample a set $\mathcal{S}_j\subseteq\mathcal{V}$ of
        $n_j$ nodes uniformly without replacement
        \State Set $t_i^j=\mathbb{I}(i\in\mathcal{S}_j)$ for all
        $i\in\mathcal{V}$
        \State Compute the predicted outcome
        \[
        \hat{y}_i^j
        =
        \hat{y}_i
        \left(
        t_i^j,
        \mathbf{t}_{\mathcal{N}_i}^j
        \right)
        \qquad
        \forall i\in\mathcal{V}
        \]
        \State Compute the ground-truth conditional mean
        \[
        \bar{y}_i^j
        =
        \bar{y}_i
        \left(
        t_i^j,
        \mathbf{t}_{\mathcal{N}_i}^j
        \right)
        \qquad
        \forall i\in\mathcal{V}
        \]
        \State Compute
        \[
        \operatorname{MSE}_j
        =
        \frac{1}{|\mathcal{V}|}
        \sum_{i\in\mathcal{V}}
        \left(
        \bar{y}_i^j
        -
        \hat{y}_i^j
        \right)^2
        \]
    \EndFor
    \State \Return
    $\displaystyle
    \operatorname{CNEE}
    =
    \frac{1}{m}
    \sum_{j=1}^{m}
    \operatorname{MSE}_j$
\end{algorithmic}
\end{algorithm}

\begin{algorithm}
\caption{PEHNE calculation}
\label{alg: PEHNE}
\begin{algorithmic}[1]
    \For{$j=1,\ldots,m$}
        \State Set
        $p_j=\frac{j-1}{m-1}$ and
        $n_j=\left\lfloor p_j|\mathcal{V}|\right\rfloor$
        \State Sample a set $\mathcal{S}_j\subseteq\mathcal{V}$ of
        $n_j$ nodes uniformly without replacement
        \State Set $t_i^j=\mathbb{I}(i\in\mathcal{S}_j)$ for all
        $i\in\mathcal{V}$
        \State Compute the estimated effect
        \[
        \hat{\omega}_i^j
        =
        \hat{y}_i
        \left(
        t_i^j,
        \mathbf{t}_{\mathcal{N}_i}^j
        \right)
        -
        \hat{y}_i(0,\mathbf{0})
        \qquad
        \forall i\in\mathcal{V}
        \]
        \State Compute the ground-truth effect
        \[
        \omega_i^j
        =
        \bar{y}_i
        \left(
        t_i^j,
        \mathbf{t}_{\mathcal{N}_i}^j
        \right)
        -
        \bar{y}_i(0,\mathbf{0})
        \qquad
        \forall i\in\mathcal{V}
        \]
        \State Compute
        \[
        \operatorname{MSE}_j
        =
        \frac{1}{|\mathcal{V}|}
        \sum_{i\in\mathcal{V}}
        \left(
        \omega_i^j
        -
        \hat{\omega}_i^j
        \right)^2
        \]
    \EndFor
    \State \Return
    $\displaystyle
    \operatorname{PEHNE}
    =
    \frac{1}{m}
    \sum_{j=1}^{m}
    \operatorname{MSE}_j$
\end{algorithmic}
\end{algorithm}

The instantiation of $\rho$ described above has two motivations. First, it evaluates the models across the full
range of treatment rates rather than only near the observed rate, where good performance
may result from interpolation around the factual assignment. Second, because treated nodes are sampled independently of covariates or network
topology, the counterfactual assignments do not preserve the treatment-configuration imbalance
induced by covariate-dependent treatment assignment. The metrics therefore also evaluate
performance on treatment configurations that are infrequent in the observed data,
the regime in which treatment-configuration imbalance requires the model
to extrapolate.
However,
$\rho$ can also be chosen to represent a treatment policy of interest, in which case CNEE and PEHNE measure estimation error under policy-relevant counterfactual assignments. 
The bound in \Cref{app:bound} applies to any 
$\rho$.

In our experiments, we set $m=50$ for both CNEE and PEHNE. 
Note that these metrics are used solely for performance evaluation, not for hyperparameter tuning: since they require all potential outcomes to be known, they cannot be calculated from observational data.

\clearpage
\section{Data-Generating Processes}\label{app: DGP}
We adapt the DGP proposed by \citet{jiang2022estimating} and \citet{caljon2025optimizing}. Instead of using a prespecified exposure mapping $z_i=\frac{1}{|\mathcal{N}_i|}\sum_{j\in \mathcal{N}_i}t_j$, we define a function that allows for heterogeneous spillover effects.

For the fully synthetic datasets, we generate networks consisting of 10,000 nodes for training, validation, and testing.
For BA Sim, we generate $d$ features for each node from a standard normal distribution: $x_i^j \sim \mathcal{N}(0, 1), j=1,\dots,d$ (following the literature, we set $d = 10$) and use the Barabási-Albert random network model \citep{barabasi1999emergence} 
to generate the network structure (we set the number of edges added per new node to 2).
For Homophily Sim, we first sample 10,000 feature vectors from a standard multivariate normal distribution and compute pairwise cosine similarities $s_{ij}$.
An edge is created when
\[
s_{ij}>\xi_{ij},
\qquad
\xi_{ij}\sim\mathcal{N}(\mu,0.025^2).
\]
Starting from $\mu=0.80$, we adjust $\mu$ until the average degree is within 0.1 of the target value 4.
Each node is then connected to its most similar node.

For the semi-synthetic datasets (Flickr, BC, and Coauthor-CS \citep{shchur2019pitfallsgraphneuralnetwork}), 
we follow \citet{jiang2022estimating} to partition each network into training, validation, and test sets using METIS \citep{karypis1998fast}. 
Following \citet{guo2020learning} and \citet{jiang2022estimating}, we then use Latent Dirichlet Allocation \citep{blei2003latent} to reduce the sparse features to a lower-dimensional representation (again setting $d=10$ following the literature).
More details on the networks for these semi-synthetic datasets are provided in \Cref{tab:datasets}.

\begin{table}[t]
    \centering
    \setlength{\tabcolsep}{3pt} 
    \begin{tabular}{lccccccccc} 
        \toprule 
        & \multicolumn{3}{c}{Flickr} & \multicolumn{3}{c}{BC} & \multicolumn{3}{c}{Coauthor-CS}\\
        \cmidrule(lr){2-4} \cmidrule(lr){5-7} \cmidrule(lr){8-10} 
        
        & Train & Validation & Test & Train & Validation & Test& Train & Validation & Test \\
        
        \midrule 
        
        \# Nodes & 2,482 & 2,461 & 2,358 & 1,716 & 1,696 & 1,784 & 6,111 & 6,111 & 6,111\\
        \# Edges & 46,268 & 14,419 & 23,529 & 17,937 & 25,408 & 14,702 &23,856 & 23,159 & 28,707 \\
        
        \bottomrule 
    \end{tabular}
    \caption{\textcolor{black}{Networks for the Flickr, BC, and Coauthor-CS datasets.}}
    \label{tab:datasets}
\end{table}

To induce the causal structure (see \Cref{fig:causal graph}), 
treatments and outcomes are simulated for the fully synthetic and the semi-synthetic datasets,
using the following parameters:
\begin{align*}
w^{XT}_j &\sim \mathrm{Unif}(-1,1), \quad j \in \{1,\dots,d\}, \\
w^{XY}_j &\sim \mathrm{Unif}(-1,1), \quad j \in \{1,\dots,d\}, \\
w^{TY}_j &\sim \mathrm{Unif}(-1,1), \quad j \in \{1,\dots,d\}, \\
w^{X_{\mathcal{N}}Y}_j &\sim \mathrm{Unif}(-1,1), \quad j \in \{1,\dots,d\}, \\
w^{T_{\mathcal{N}}Y}_j &\sim \mathrm{Unif}(-1,1), \quad j \in \{1,\dots,d\},
\end{align*}
\vskip -0.15in
\begin{align*}
\mathbf{w}^{XT} &= [w^{XT}_1,\dots,w^{XT}_d], \\
\mathbf{w}^{XY} &= [w^{XY}_1,\dots,w^{XY}_d], \\
\mathbf{w}^{TY} &= [w^{TY}_1,\dots,w^{TY}_d], \\
\mathbf{w}^{X_{\mathcal{N}}Y} &= [w^{X_{\mathcal{N}}Y}_1,\dots,w^{X_{\mathcal{N}}Y}_d], \\
\mathbf{w}^{T_{\mathcal{N}}Y} &= [w^{T_{\mathcal{N}}Y}_1,\dots,w^{T_{\mathcal{N}}Y}_d].
\end{align*}
These parameters influence the effect of $\mathbf{X}_i$ on ${T_i}$, $\mathbf{X}_i$ on $Y_i$, the heterogeneous effect of ${T_i}$ on $Y_i$, the effect of $\mathbf{X}_{\mathcal{N}_i}$ on $Y_i$, and the heterogeneous spillover effect of $\mathbf{T}_{\mathcal{N}_i}$ on $Y_i$, respectively. 

Treatment $t_i$ is generated as follows. We first calculate $\nu_i$:
\begin{equation*}
    \nu_i= \beta_{XT}\cdot\mathbf{w}^{XT}\cdot \mathbf{x}_i,
\end{equation*}
with $\beta_{XT} \geq 0$ 
controlling the strength of covariate-dependent treatment assignment,
and $\mathbf{x}_i = [x_i^1, x_i^2, \dots, x_i^{d}]^\prime$.
Next, to set the percentage of nodes treated to approximately 25\%, we calculate the 75th percentile $\nu_{75}$ and transform $\nu^\prime=\nu-\nu_{75}$.
Finally, we apply the sigmoid function $\sigma$ to $\nu^\prime$, and obtain $t_i$ by sampling: 
\begin{equation*}
    t_i \sim \text{Bernoulli}(\sigma(\nu^\prime_{i})).
\end{equation*}
Together, the construction of $\nu^\prime$ and the sigmoid transformation place the 75th-percentile node at propensity 0.5, 
but do not constrain the realized treatment rate exactly.
When $\beta_{XT}=0$, we deviate from this procedure and sample exactly 25\% of the nodes uniformly without replacement.
Unless explicitly stated otherwise, we set $\beta_{XT} = 6$.

Outcome $y_i$ is generated as follows. 
We first create a transformed feature vector $\tilde{\mathbf{x}}_i$ 
by applying the sigmoid function $\sigma$ to half of the features to introduce nonlinearities.
$y_i$ is then obtained as follows:
\begin{equation*}
y_i =  \beta_{{\text{individual}}}\cdot h_i \cdot t_i + \beta_{{\text{spillover}}} \cdot z_i + \beta_{XY} \cdot u_i
+ \beta_{X_\mathcal{N}Y} \cdot u_{\mathcal{N}_i} + \beta_{\epsilon} \cdot \epsilon;\qquad
\epsilon \sim \mathcal{N}(0,1),
\end{equation*}
with
\begin{align*}
    h_i &= \mathbf{w}^{TY} \cdot \tilde{\mathbf{x}}_i,\\
    z_i &= \frac{1}{|\mathcal{N}_i|} \sum_{j\in\mathcal{N}_i}t_j\cdot\mathbf{w}^{T_{\mathcal{N}}Y}\cdot \tilde{\mathbf{x}}_j, \\
    u_i &= \mathbf{w}^{XY}\cdot \tilde{\mathbf{x}}_i, \\
    u_{\mathcal{N}_i} &= \frac{1}{|\mathcal{N}_i|}\sum_{j\in\mathcal{N}_i}\mathbf{w}^{X_{\mathcal{N}}Y}\cdot \tilde{\mathbf{x}}_j.
\end{align*}
Unless explicitly stated otherwise, we set $\beta_{{\text{individual}}} = 2, \beta_{\text{spillover}}=2, \beta_{XY} = 1.5, \beta_{X_\mathcal{N}Y} = 1.5$, and $\beta_{\epsilon} = 0.2$.

\clearpage
\section{Hyperparameter Selection and Implementation Details}\label{app:Hyperparam selection}
Due to the fundamental problem of causal inference \citep{holland1986statistics}, individualized treatment effects are unobservable. As a result, selecting hyperparameters is challenging, since we cannot directly optimize based on treatment effect estimation error. 
For standard machine learning hyperparameters, such as hidden layer size or learning rate, we can rely on the factual validation loss for selection. The factual validation loss is the average estimation error for outcomes actually observed in the validation set and can always be calculated. However, the factual loss may not reflect the treatment effect estimation performance. Nevertheless, this approach has been shown to work reasonably well \citep{curth2023search}. 

The weight for adversarial balancing $\alpha$ is a special type of hyperparameter. A positive $\alpha$ may cause the model to discard relevant information for predicting the observed outcomes in favor of constructing treatment-invariant representations, which will likely increase the factual validation loss. Consequently, if the factual loss is used to select this hyperparameter, $\alpha$ will often be set to zero, meaning that the treatment-prediction branch of HINet would not be used. 
However, both theoretical and empirical work suggests that balancing representations can improve treatment effect estimates \citep{shalit2017estimating,bica2020estimating,berrevoets2020}.
Motivated by this, we propose the following approach for selecting $\alpha$. First, the standard machine learning hyperparameters are tuned using the factual validation loss. Once these hyperparameters are set, the factual loss is calculated for different values of $\alpha$. As $\alpha$ increases, the factual loss typically increases as well. A modest increase is acceptable, whereas a substantial increase may indicate that outcome-relevant information is being discarded in favor of learning treatment-invariant representations. As a heuristic, we therefore select the largest value of $\alpha$ for which the factual loss remains below $(1+p)\cdot\text{loss}_{\alpha=0}$, with tolerance $p = 0.10$. An important advantage of this approach is that it allows $\alpha=0$ to be selected when representation balancing would otherwise result in excessive information being discarded. This safeguard depends on the factual loss reacting to $\alpha$. In \Cref{sec:failure_modes_heuristic},
we report
two DGPs for which the factual loss reacts only weakly to changes in $\alpha$, so that the tolerance results in selecting an $\alpha$ for which counterfactual estimation error has already increased substantially.
 
$\alpha$ is selected from the range $\{0,0.025,0.05,0.1,0.2,0.3\}$ for HINet and NetEst. 
The other hyperparameters for these methods and for TNet are selected from the ranges shown in \Cref{tab:hyperparams}.

\begin{table}[htbp]
    \centering
    \begin{tabular}{lc} 
        Parameter& Value\\
        \midrule
        Hidden size & $\{16,32\}$\\
        Num. epochs  & $\{500,1000,2000\}$\\
        Initial learning rate  & $\{0.001,0.0005,0.0001\}$\\
        Dropout probability  & $\{0.0,0.1,0.2\}$\\
        \bottomrule 
    \end{tabular}
    \caption{Hyperparameter ranges for all methods.}
    \label{tab:hyperparams}
\end{table}

All GIN layers internally use a 2-layer MLP and have $\epsilon=0$. The encoder block $e_\phi$ in HINet consists of two hidden layers, whereas the MLP blocks $d_T$ and $p_Y$, as well as the MLP block in the GIN model, each consist of three hidden layers. All MLP blocks (for every method) use ReLU activations after each layer. Other hyperparameters are set to author-recommended values. 

Each model is trained using the Adam optimizer \citep{KingBa15} with weight decay set to $0.001$. For all models except SPNet, TNet, and IDE-Net, we use the implementation provided by \citet{jiang2022estimating}. We implemented SPNet ourselves based on the description in \citet{huang2023modeling}, 
as no publicly available implementation exists. For TNet, we used the implementation from \citet{chendoubly} and set $\alpha=\gamma=1$, as this yielded stable results in terms of factual validation loss for all datasets. For IDE-Net, we adapt the authors' original implementation \citep{adhikari2025inferring}, retaining its heterogeneous
exposure module (a two-layer exposure mapper over neighbors' treatments and
mapped features), its original feature mapper (two layers, with a binary edge
indicator as edge feature), its variance-smoothing regularizer (weight $0.1$,
enabled after a 150-epoch warm-up), early stopping (patience 300), an
estimator learning rate of $10\times$ the representation learning rate with
step decay (halved every 50 epochs), and gradient-norm clipping at $3$. The above hyperparameters are the same as those used in the original paper. Its
standard hyperparameters are tuned using the same ranges as all other methods.

All reported results are averages over five different initializations, affecting both weight initialization and training data shuffling. 

\textbf{Reproducibility.}\; Our code is available at \href{https://github.com/daan-caljon/HINet}{https://github.com/daan-caljon/HINet}. All experiments were run on Linux using an Intel Xeon Gold 6140 CPU, 45 GiB of RAM, and an NVIDIA P100 GPU.

\FloatBarrier

\clearpage
\section{Performance under Other Exposure Mappings}\label{app:other_mappings}
To examine how the methods perform under different exposure mappings, we repeat the main experiment using several alternative exposure mappings in the DGP.

\Cref{tab:results_exp_sum} reports the test set results for the \emph{sum} of treatments of neighbors as 
the exposure mapping, i.e., $z_i = \sum_{j\in \mathcal{N}_i}t_j$. The results show that both NetEst and TNet perform poorly because their assumed exposure mappings differ substantially from the true one.
This highlights how misspecification of the exposure mapping can lead to inaccurate treatment effect estimates. In contrast, HINet and the GIN model---neither of which requires a prespecified exposure mapping---achieve the best performance. 

\Cref{tab:results_exp_avg} reports the test set results for the \emph{proportion} of
treated neighbors as the exposure mapping, i.e.,
$z_i=\frac{1}{|\mathcal{N}_i|}\sum_{j\in\mathcal{N}_i} t_j$, which is the
exposure mapping assumed by NetEst and TNet. 
However, even under this DGP, HINet performs best on seven of the ten dataset--metric combinations.
TNet performs best on PEHNE for BA Sim, Homophily Sim, and Coauthor-CS, with HINet ranking second in each case. 
A possible explanation for this is that the GCN aggregation of NetEst and TNet cannot capture the nonlinear dependence of $Y_i$ on $\mathbf{X}_{\mathcal{N}_i}$ present in the DGP. Supporting this explanation, TNet becomes the best-performing method once this dependence is removed (see \Cref{tab:avg_without}).

To provide further evidence that HINet can learn a wide variety of exposure
mappings, we report results for two additional mappings. In
\Cref{tab:results_entropy}, we use \emph{information entropy} of treatments of neighbors as the exposure
mapping: $z_i=-p\log_2(p)-(1-p)\log_2(1-p)-0.5$, where $p$ is the proportion
of treated neighbors (we subtract $0.5$ to allow for negative spillover
effects). In \Cref{tab:results_squared}, we use the 
\emph{squared feature-weighted average} of treatments
of neighbors as the exposure mapping:
$z_i=\frac{1}{|\mathcal{N}_i|}\sum_{j\in\mathcal{N}_i} w^2(\mathbf{x}_j)\,t_j$.
Under the squared feature-weighted average mapping, HINet has the lowest reported mean error on all dataset–metric combinations except PEHNE on Coauthor-CS, where TNet is marginally ahead. Under the information entropy mapping, HINet performs best on most combinations and second-best on the remaining ones: the GIN model has the lowest CNEE on BC and the lowest error on Flickr for both metrics, and TNet has the lowest PEHNE on Homophily Sim. In each of these cases the difference is small relative to the variation across initializations, so we do not interpret it as a performance gap. The GIN model is the closest competitor on BC and Flickr under both exposure mappings, which is consistent with the ability of GINs to represent different forms of neighborhood exposure without a prespecified mapping. On Coauthor-CS under the information entropy mapping, TNet has a much larger error with a comparably large standard deviation, indicating that its estimates vary strongly across initializations on this dataset. 

Across all five exposure mappings considered in the DGP (the four mappings reported here and the default mapping used for \Cref{tab:results}),
the best competing method changes from mapping to mapping, while HINet is
consistently best or near-best. HINet's consistency is most pronounced on Coauthor-CS, where it has the lowest CNEE under every mapping. On PEHNE, its average rank on this dataset (1.4) is likewise the best, ahead of IDE-Net (3.2) and TNet (3.4).
TNet is the strongest competitor when its
assumed proportion mapping is a reasonable approximation. However, on the sum and information entropy mappings it performs poorly compared to HINet and the GIN model.

To examine the impact of the direct influence of $\mathbf{X}_{\mathcal{N}_i}$ on $Y_i$ in the DGP,
we repeat the experiment with this influence removed (i.e., we set $\beta_{X_{\mathcal{N}}Y}=0$), 
under both the
proportion and the (default) feature-weighted average exposure mappings.
\Cref{tab:avg_without,tab:weight_without} report the respective test set results.
Consistent with the explanation offered for \Cref{tab:results_exp_avg}, TNet is the
most accurate method under the proportion mapping, where its assumed exposure
mapping is correctly specified. Under the feature-weighted average mapping, TNet is best
only on Coauthor-CS, with TARNet and the GIN model being the most accurate on the
remaining datasets. 

The capacity to capture the direct influence of $\mathbf{X}_{\mathcal{N}_i}$ on $Y_i$ 
accounts for part of HINet's
strength: with this direct influence removed from the DGP,
the test set results in \Cref{tab:alpha_scope} for HINet without balancing ($\alpha=0$) show that it 
performs best on BC under the proportion mapping and is otherwise competitive with,
but not necessarily better than, the strongest competitors. HINet with the heuristically selected $\alpha$ is less accurate on both mappings, which reflects the impact of 
the selected $\alpha$ rather than an inferior outcome model. 
The next section
examines why the selection heuristic discussed in \Cref{app:Hyperparam selection} selects harmful values of $\alpha$ for these two DGPs.

\begin{table}[!t]
\centering
\begin{adjustbox}{width=\textwidth}
\begin{tabular}{llcccccccc}
\toprule
Dataset & Metric & TARNet & NetDeconf & NetEst & TNet & GIN model & SPNet & IDE-Net & HINet (ours) \\
\midrule
\multirow{2}{*}{BC} & PEHNE & 599.64 $\pm$ 10.84 & 637.36 $\pm$ 6.92 & 298.82 $\pm$ 21.04 & 606.56 $\pm$ 49.55 & \underline{27.93 $\pm$ 2.76} & 553.16 $\pm$ 3.75 & 272.10 $\pm$ 33.93 & \textbf{21.76 $\pm$ 1.57} \\
 & CNEE & 373.18 $\pm$ 6.28 & 364.09 $\pm$ 5.17 & 311.16 $\pm$ 11.67 & 511.80 $\pm$ 54.01 & \underline{24.56 $\pm$ 2.38} & 260.43 $\pm$ 2.56 & 172.53 $\pm$ 20.59 & \textbf{19.89 $\pm$ 1.52} \\
\midrule
\multirow{2}{*}{Flickr} & PEHNE & 2837.76 $\pm$ 20.98 & 2872.94 $\pm$ 16.29 & 2868.29 $\pm$ 29.25 & 2781.49 $\pm$ 49.09 & \textbf{369.57 $\pm$ 39.80} & 2706.12 $\pm$ 62.11 & 2468.06 $\pm$ 159.01 & \underline{494.85 $\pm$ 165.23} \\
 & CNEE & 2503.23 $\pm$ 13.31 & 2478.42 $\pm$ 10.23 & 2765.32 $\pm$ 33.51 & 2777.92 $\pm$ 25.38 & \textbf{310.00 $\pm$ 21.47} & 1077.76 $\pm$ 50.63 & 1069.69 $\pm$ 66.36 & \underline{474.38 $\pm$ 156.37} \\
\midrule
\multirow{2}{*}{BA Sim} & PEHNE & 77.56 $\pm$ 0.65 & 77.97 $\pm$ 0.48 & 59.19 $\pm$ 0.46 & 60.27 $\pm$ 2.58 & \underline{11.87 $\pm$ 0.52} & 79.61 $\pm$ 2.13 & 51.73 $\pm$ 1.08 & \textbf{10.10 $\pm$ 0.69} \\
 & CNEE & 68.06 $\pm$ 0.38 & 69.00 $\pm$ 0.18 & 58.42 $\pm$ 0.57 & 58.68 $\pm$ 1.29 & \underline{11.36 $\pm$ 0.42} & 40.90 $\pm$ 1.14 & 30.93 $\pm$ 0.55 & \textbf{9.68 $\pm$ 0.62} \\
\midrule
\multirow{2}{*}{Homophily Sim} & PEHNE & 27.37 $\pm$ 0.60 & 31.01 $\pm$ 0.33 & 6.39 $\pm$ 0.22 & 7.88 $\pm$ 0.53 & \underline{0.36 $\pm$ 0.08} & 23.52 $\pm$ 0.64 & 1.65 $\pm$ 0.15 & \textbf{0.22 $\pm$ 0.03} \\
 & CNEE & 21.62 $\pm$ 0.58 & 24.37 $\pm$ 0.36 & 6.24 $\pm$ 0.19 & 7.61 $\pm$ 0.54 & \underline{0.41 $\pm$ 0.05} & 17.63 $\pm$ 0.56 & 1.56 $\pm$ 0.17 & \textbf{0.25 $\pm$ 0.04} \\
 \midrule
\multirow{2}{*}{Coauthor-CS} & PEHNE & 207.68 $\pm$ 0.75 & 206.53 $\pm$ 6.11 & 208.34 $\pm$ 8.59 & 160.55 $\pm$ 3.71 & \underline{46.35 $\pm$ 5.70} & 185.67 $\pm$ 1.50 & 70.52 $\pm$ 9.58 & \textbf{5.68 $\pm$ 0.49} \\
 & CNEE & 171.15 $\pm$ 1.19 & 172.25 $\pm$ 4.67 & 174.39 $\pm$ 6.98 & 169.40 $\pm$ 4.67 & \underline{38.27 $\pm$ 5.37} & 145.42 $\pm$ 2.01 & 48.77 $\pm$ 6.22 & \textbf{5.33 $\pm$ 0.47} \\
\bottomrule
\end{tabular}
\end{adjustbox}
\caption{Test set results (mean $\pm$ SD over five different initializations) for the \emph{sum} of neighbors' treatments used as exposure mapping in the DGP. Lower is better for both metrics. The best-performing method is in bold; the second-best is underlined.}
\label{tab:results_exp_sum}
\end{table}

\begin{table}[!t]
\centering
\begin{adjustbox}{width=\textwidth}
\begin{tabular}{llcccccccc}
\toprule
Dataset & Metric & TARNet & NetDeconf & NetEst & TNet & GIN model & SPNet & IDE-Net & HINet (ours) \\
\midrule
\multirow{2}{*}{BC} & PEHNE & 1.68 $\pm$ 0.22 & 4.58 $\pm$ 0.10 & 1.97 $\pm$ 0.13 & \underline{1.08 $\pm$ 0.07} & 1.09 $\pm$ 0.07 & 4.37 $\pm$ 0.07 & 1.44 $\pm$ 0.17 & \textbf{0.66 $\pm$ 0.18} \\
 & CNEE & 1.94 $\pm$ 0.23 & 4.50 $\pm$ 0.09 & 1.94 $\pm$ 0.09 & 1.05 $\pm$ 0.07 & \underline{0.95 $\pm$ 0.08} & 4.38 $\pm$ 0.04 & 1.43 $\pm$ 0.20 & \textbf{0.66 $\pm$ 0.21} \\
\midrule
\multirow{2}{*}{Flickr} & PEHNE & 1.64 $\pm$ 0.16 & 4.73 $\pm$ 0.52 & 2.50 $\pm$ 0.09 & 1.60 $\pm$ 0.28 & \underline{1.32 $\pm$ 0.13} & 5.19 $\pm$ 0.23 & 2.16 $\pm$ 0.39 & \textbf{0.54 $\pm$ 0.09} \\
 & CNEE & 2.33 $\pm$ 0.17 & 5.94 $\pm$ 0.42 & 3.77 $\pm$ 0.11 & 2.12 $\pm$ 0.29 & \underline{1.30 $\pm$ 0.14} & 6.67 $\pm$ 0.25 & 2.58 $\pm$ 0.53 & \textbf{0.63 $\pm$ 0.10} \\
\midrule
\multirow{2}{*}{BA Sim} & PEHNE & 1.42 $\pm$ 0.05 & 3.63 $\pm$ 0.01 & 0.49 $\pm$ 0.01 & \textbf{0.16 $\pm$ 0.01} & 1.12 $\pm$ 0.09 & 4.19 $\pm$ 0.32 & 0.38 $\pm$ 0.06 & \underline{0.24 $\pm$ 0.07} \\
 & CNEE & 2.66 $\pm$ 0.02 & 5.10 $\pm$ 0.01 & 1.12 $\pm$ 0.04 & 0.62 $\pm$ 0.01 & 0.94 $\pm$ 0.09 & 5.48 $\pm$ 0.34 & \underline{0.42 $\pm$ 0.06} & \textbf{0.27 $\pm$ 0.08} \\
\midrule
\multirow{2}{*}{Homophily Sim} & PEHNE & 2.16 $\pm$ 0.30 & 1.26 $\pm$ 0.03 & 0.30 $\pm$ 0.09 & \textbf{0.09 $\pm$ 0.01} & 0.26 $\pm$ 0.03 & 1.79 $\pm$ 0.07 & 0.38 $\pm$ 0.07 & \underline{0.17 $\pm$ 0.04} \\
 & CNEE & 2.25 $\pm$ 0.30 & 1.18 $\pm$ 0.02 & 0.46 $\pm$ 0.07 & \underline{0.22 $\pm$ 0.01} & 0.34 $\pm$ 0.03 & 1.67 $\pm$ 0.06 & 0.44 $\pm$ 0.07 & \textbf{0.19 $\pm$ 0.04} \\
 \midrule
\multirow{2}{*}{Coauthor-CS} & PEHNE & 2.47 $\pm$ 0.32 & 3.36 $\pm$ 0.39 & 2.36 $\pm$ 0.15 & \textbf{0.69 $\pm$ 0.16} & 2.09 $\pm$ 0.14 & 3.92 $\pm$ 1.00 & 1.78 $\pm$ 0.30 & \underline{0.84 $\pm$ 0.18} \\
 & CNEE & 2.34 $\pm$ 0.18 & 2.61 $\pm$ 0.22 & 2.20 $\pm$ 0.21 & \underline{0.98 $\pm$ 0.12} & 2.05 $\pm$ 0.18 & 3.29 $\pm$ 0.56 & 1.83 $\pm$ 0.34 & \textbf{0.92 $\pm$ 0.17} \\
\bottomrule
\end{tabular}
\end{adjustbox}
\caption{Test set results (mean $\pm$ SD over five different initializations) for the \emph{proportion} of treated neighbors used as exposure mapping in the DGP. Lower is better for both metrics. The best-performing method is in bold; the second-best is underlined.}
\label{tab:results_exp_avg}
\end{table}

\begin{table}[!t]
\centering
\begin{adjustbox}{width=\textwidth}
\begin{tabular}{llcccccccc}
\toprule
Dataset & Metric & TARNet & NetDeconf & NetEst & TNet & GIN model & SPNet & IDE-Net & HINet (ours) \\
\midrule
\multirow{2}{*}{BC} & PEHNE & 2.11 $\pm$ 0.20 & 5.02 $\pm$ 0.09 & 2.26 $\pm$ 0.21 & 3.41 $\pm$ 1.70 & \underline{1.19 $\pm$ 0.09} & 5.88 $\pm$ 0.57 & 1.72 $\pm$ 0.18 & \textbf{1.04 $\pm$ 0.25} \\
 & CNEE & 2.07 $\pm$ 0.20 & 5.46 $\pm$ 0.08 & 2.60 $\pm$ 0.23 & 3.22 $\pm$ 1.71 & \textbf{1.17 $\pm$ 0.08} & 6.41 $\pm$ 0.66 & 1.89 $\pm$ 0.30 & \underline{1.21 $\pm$ 0.35} \\
\midrule
\multirow{2}{*}{Flickr} & PEHNE & 1.51 $\pm$ 0.12 & 5.25 $\pm$ 0.31 & 2.67 $\pm$ 0.18 & 1.33 $\pm$ 0.34 & \textbf{1.10 $\pm$ 0.06} & 6.00 $\pm$ 0.21 & 3.22 $\pm$ 0.61 & \underline{1.11 $\pm$ 0.38} \\
 & CNEE & 2.37 $\pm$ 0.20 & 7.66 $\pm$ 0.34 & 4.09 $\pm$ 0.23 & 1.99 $\pm$ 0.32 & \textbf{1.09 $\pm$ 0.03} & 8.65 $\pm$ 0.20 & 3.92 $\pm$ 0.62 & \underline{1.30 $\pm$ 0.49} \\
\midrule
\multirow{2}{*}{BA Sim} & PEHNE & 1.20 $\pm$ 0.04 & 3.86 $\pm$ 0.02 & 0.56 $\pm$ 0.04 & \underline{0.24 $\pm$ 0.06} & 1.41 $\pm$ 0.14 & 4.28 $\pm$ 0.42 & 0.61 $\pm$ 0.12 & \textbf{0.21 $\pm$ 0.07} \\
 & CNEE & 2.66 $\pm$ 0.08 & 5.88 $\pm$ 0.02 & 1.12 $\pm$ 0.04 & 0.67 $\pm$ 0.07 & 1.22 $\pm$ 0.14 & 6.09 $\pm$ 0.40 & \underline{0.63 $\pm$ 0.12} & \textbf{0.24 $\pm$ 0.08} \\
\midrule
\multirow{2}{*}{Homophily Sim} & PEHNE & 1.58 $\pm$ 0.06 & 1.68 $\pm$ 0.01 & 0.36 $\pm$ 0.08 & \textbf{0.13 $\pm$ 0.04} & 0.39 $\pm$ 0.02 & 1.67 $\pm$ 0.06 & 0.51 $\pm$ 0.11 & \underline{0.14 $\pm$ 0.08} \\
 & CNEE & 1.58 $\pm$ 0.06 & 1.48 $\pm$ 0.01 & 0.51 $\pm$ 0.07 & \underline{0.27 $\pm$ 0.04} & 0.47 $\pm$ 0.03 & 1.46 $\pm$ 0.07 & 0.55 $\pm$ 0.11 & \textbf{0.16 $\pm$ 0.09} \\
 \midrule
\multirow{2}{*}{Coauthor-CS} & PEHNE & 1.98 $\pm$ 0.24 & 3.47 $\pm$ 0.42 & \underline{1.13 $\pm$ 0.08} & 12326.22 $\pm$ 20587.49 & 2.94 $\pm$ 0.24 & 3.75 $\pm$ 0.79 & 1.26 $\pm$ 0.11 & \textbf{1.01 $\pm$ 0.30} \\
 & CNEE & 1.76 $\pm$ 0.13 & 2.61 $\pm$ 0.24 & 1.77 $\pm$ 0.41 & 10271.75 $\pm$ 16366.98 & 3.22 $\pm$ 0.31 & 3.11 $\pm$ 0.38 & \underline{1.36 $\pm$ 0.07} & \textbf{1.11 $\pm$ 0.37} \\
\bottomrule
\end{tabular}
\end{adjustbox}
\caption{Test set results (mean $\pm$ SD over five different initializations) for the \emph{information entropy} of neighbors' treatments used as exposure mapping in the DGP. Lower is better for both metrics. The best-performing method is in bold; the second-best is underlined.}
\label{tab:results_entropy}
\end{table}

\begin{table}[!t]
\centering
\begin{adjustbox}{width=\textwidth}
\begin{tabular}{llcccccccc}
\toprule
Dataset & Metric & TARNet & NetDeconf & NetEst & TNet & GIN model & SPNet & IDE-Net & HINet (ours) \\
\midrule
\multirow{2}{*}{BC} & PEHNE & 3.16 $\pm$ 0.05 & 4.61 $\pm$ 0.05 & 2.14 $\pm$ 0.11 & 1.42 $\pm$ 0.24 & \underline{1.00 $\pm$ 0.07} & 4.37 $\pm$ 0.10 & 1.40 $\pm$ 0.28 & \textbf{0.50 $\pm$ 0.07} \\
 & CNEE & 3.85 $\pm$ 0.07 & 4.78 $\pm$ 0.02 & 2.09 $\pm$ 0.09 & 1.45 $\pm$ 0.24 & \underline{0.85 $\pm$ 0.08} & 4.49 $\pm$ 0.13 & 1.38 $\pm$ 0.30 & \textbf{0.51 $\pm$ 0.07} \\
\midrule
\multirow{2}{*}{Flickr} & PEHNE & 3.60 $\pm$ 0.03 & 4.55 $\pm$ 0.13 & 3.08 $\pm$ 0.18 & 1.62 $\pm$ 0.12 & \underline{1.35 $\pm$ 0.09} & 5.63 $\pm$ 0.32 & 2.55 $\pm$ 0.66 & \textbf{0.86 $\pm$ 0.23} \\
 & CNEE & 5.34 $\pm$ 0.03 & 6.19 $\pm$ 0.15 & 4.47 $\pm$ 0.24 & 2.45 $\pm$ 0.06 & \underline{1.41 $\pm$ 0.11} & 7.41 $\pm$ 0.35 & 2.94 $\pm$ 0.70 & \textbf{1.00 $\pm$ 0.26} \\
\midrule
\multirow{2}{*}{BA Sim} & PEHNE & 3.77 $\pm$ 0.01 & 4.21 $\pm$ 0.01 & 1.40 $\pm$ 0.09 & \underline{1.03 $\pm$ 0.02} & 1.77 $\pm$ 0.02 & 4.79 $\pm$ 0.36 & 1.30 $\pm$ 0.12 & \textbf{0.34 $\pm$ 0.02} \\
 & CNEE & 5.47 $\pm$ 0.02 & 5.96 $\pm$ 0.02 & 2.17 $\pm$ 0.08 & 1.51 $\pm$ 0.03 & 1.70 $\pm$ 0.03 & 6.32 $\pm$ 0.36 & \underline{1.28 $\pm$ 0.12} & \textbf{0.33 $\pm$ 0.02} \\
\midrule
\multirow{2}{*}{Homophily Sim} & PEHNE & 4.06 $\pm$ 0.04 & 3.13 $\pm$ 0.40 & \underline{1.02 $\pm$ 0.07} & 3.22 $\pm$ 2.75 & 1.03 $\pm$ 0.08 & 2.51 $\pm$ 0.06 & 1.17 $\pm$ 0.17 & \textbf{0.29 $\pm$ 0.11} \\
 & CNEE & 4.37 $\pm$ 0.03 & 2.86 $\pm$ 0.40 & 1.18 $\pm$ 0.08 & 3.14 $\pm$ 2.85 & \underline{1.12 $\pm$ 0.10} & 2.29 $\pm$ 0.06 & 1.15 $\pm$ 0.17 & \textbf{0.30 $\pm$ 0.12} \\
 \midrule
\multirow{2}{*}{Coauthor-CS} & PEHNE & 4.80 $\pm$ 0.08 & 3.83 $\pm$ 0.22 & 4.38 $\pm$ 0.16 & \textbf{2.89 $\pm$ 0.15} & 3.41 $\pm$ 0.13 & 5.26 $\pm$ 1.11 & {3.12 $\pm$ 0.23} & \underline{2.90 $\pm$ 0.27} \\
 & CNEE & 5.50 $\pm$ 0.10 & 4.13 $\pm$ 0.17 & 4.48 $\pm$ 0.26 & \underline{3.69 $\pm$ 0.22} & 4.53 $\pm$ 0.19 & 5.23 $\pm$ 0.59 & 3.91 $\pm$ 0.09 & \textbf{3.00 $\pm$ 0.19} \\

\bottomrule
\end{tabular}
\end{adjustbox}
\caption{Test set results (mean $\pm$ SD over five different initializations) for the \emph{squared feature-weighted average} of neighbors' treatments used as exposure mapping in the DGP. Lower is better for both metrics. The best-performing method is in bold; the second-best is underlined.}
\label{tab:results_squared}
\end{table}
\FloatBarrier

\clearpage
\section{Failure Modes of Factual-Loss-Based Selection of \texorpdfstring{$\alpha$}{alpha}} \label{sec:failure_modes_heuristic}

The $\alpha$-selection heuristic selects the largest $\alpha$ for
which the factual validation loss remains within a factor $(1 + p)$ of its value
at $\alpha=0$, and therefore constrains the factual cost of balancing but not the
counterfactual cost. \Cref{tab:alpha_scope} compares
the test set results for
HINet under the selected $\alpha$ with the
no-balancing variant ($\alpha = 0$)
for two DGPs (the proportion and feature-weighted average exposure mappings) without direct influence from $\mathbf{X}_{\mathcal{N}_i}$ on $Y_i$ (i.e., with $\beta_{X_{\mathcal{N}}Y}=0$; also see \Cref{tab:avg_without,tab:weight_without}), and reports the factual validation loss alongside test PEHNE and CNEE. On
all datasets except Coauthor-CS, the heuristic allows for substantial balancing: the
factual loss at the selected $\alpha$ stays within the tolerance, while CNEE and PEHNE increase substantially.

How loose the constraint is depends on the scale of the factual loss, because the
tolerance is multiplicative. The irreducible noise variance in our DGP is 0.04,
making 0.04 the lowest attainable factual loss. On all datasets
except Coauthor-CS, the factual loss at $\alpha= 0$ lies relatively close to this floor. When the factual loss is close to 0.04, most of its value is irreducible and cannot be affected by $\alpha$ at all. A tolerance of
$p = 0.10$ can therefore allow for a much larger proportional increase in the reducible loss (the difference between the total and the irreducible factual loss). On Coauthor-CS, the factual loss is an order of magnitude
above the noise floor. On this dataset, the heuristic works as intended and permits an increase of only about a tenth in the reducible part,
selecting $\alpha=0$ under the proportion mapping and the smallest positive
candidate under the feature-weighted average mapping. The effective constraint imposed by $p$ becomes looser as the factual loss approaches the noise floor. Consequently, the heuristic can select a large $\alpha$ even when balancing does not improve counterfactual accuracy.

Under these two DGP variants, balancing may also be comparatively inexpensive in factual terms because the direct neighborhood-covariate term has been removed. Under
the proportion mapping, the potential outcomes do not depend on $\mathbf{X}_{\mathcal{N}_i}$, so strong ignorability holds conditional on $\mathbf{X}_i$ alone. The neighborhood component of treatment-configuration imbalance therefore no longer involves outcome-relevant neighbor covariates. Under the feature-weighted average mapping, the neighbors' covariates affect the outcome only when those neighbors are treated.
Consequently, under both mappings,
and particularly under the proportion mapping, removing some neighborhood information may have little effect on factual prediction. The factual validation loss can then remain within the selection tolerance even when counterfactual accuracy deteriorates. However, the node's own covariates remain outcome-relevant, so an explanation for why degrading their representation might also be factually cheap is needed: under strong covariate-dependent treatment assignment, the observed treatments are informative of the covariate information that balancing removes, so the outcome model can partly substitute one for the other. This substitution only works when treatments and covariates are associated,
which they are in the DGPs used. CNEE and PEHNE are computed over counterfactual networks in which treatments are assigned independently of covariates, so the association is absent and this substitution no longer works.
The feature-weighted average exposure mapping keeps a factual role for the treated neighbors' covariates, so less of the discarded information can be recovered through substitution. Therefore, balancing becomes more expensive in factual terms under this mapping, so the tolerance is hit sooner, and we would expect a smaller increase in error at the selected $\alpha$ than under the proportion mapping. This is what we find on all four datasets where the heuristic
selects $\alpha>0$ under both mappings. The observed pattern is consistent with representation-induced confounding: the selected $\alpha$ may remove outcome-relevant covariate information without producing a substantial increase in factual loss \citep{melnychuk2024bounds}.
The factual validation loss cannot reveal this, since under the observed treatment assignment the treatments substitute for the information that was removed.

The heuristic is informative only when the removal of outcome-relevant information produces a detectable increase in factual validation loss. Under the default DGP, in which the outcome depends on the neighbors' covariates both directly and through the exposure mapping, HINet with the heuristically selected $\alpha$ is comparable to or more accurate than its unbalanced ($\alpha=0$) variant (see \Cref{fig:cnee_rep_bal,fig:sensitivity}).

A nearly flat factual-validation-loss curve across candidate values of $\alpha$ is a practical warning sign, because the observed loss then provides little signal for distinguishing among them. When the variation across candidates is small, the evidence for choosing $\alpha>0$ is weak, and $\alpha=0$ is the conservative choice. However, the presence of variation is not by itself sufficient to justify balancing. 
The fact that \emph{no observable criterion guarantees counterfactual accuracy is not specific to our heuristic} and is a well-known problem in causal machine learning \citep{curth2023search}. A treatment-configuration-aware selection criterion would require estimating or approximating the propensity $p(\mathbf{A}_i | \mathbf{C}_i)$ over treatment configurations, which is impractical without making additional assumptions. We therefore use the factual-loss-based heuristic as a practical, observable constraint on the cost of balancing and examine sensitivity to both $\alpha$ and $p$ in \Cref{fig:sensitivity}.

\begin{table}[t]
\centering
\begin{adjustbox}{width=\textwidth}
\begin{tabular}{llcccccccc}
\toprule
Dataset & Metric & TARNet & NetDeconf & NetEst & TNet & GIN model & SPNet & IDE-Net & HINet (ours) \\
\midrule
\multirow{2}{*}{BC} & PEHNE & 1.58 $\pm$ 0.09 & 4.55 $\pm$ 0.15 & 1.77 $\pm$ 0.11 & \underline{1.14 $\pm$ 0.78} & \textbf{0.78 $\pm$ 0.04} & 4.89 $\pm$ 0.31 & 1.21 $\pm$ 0.28 & 2.27 $\pm$ 0.28 \\
 & CNEE & \underline{0.90 $\pm$ 0.07} & 4.11 $\pm$ 0.13 & 1.75 $\pm$ 0.11 & 1.11 $\pm$ 0.71 & \textbf{0.72 $\pm$ 0.04} & 4.53 $\pm$ 0.30 & 1.21 $\pm$ 0.30 & 2.30 $\pm$ 0.29 \\
\midrule
\multirow{2}{*}{Flickr} & PEHNE & 1.54 $\pm$ 0.07 & 4.51 $\pm$ 0.35 & 1.89 $\pm$ 0.57 & \textbf{0.36 $\pm$ 0.08} & \underline{1.02 $\pm$ 0.11} & 5.01 $\pm$ 0.21 & 2.11 $\pm$ 0.21 & 3.47 $\pm$ 0.27 \\
 & CNEE & \underline{0.87 $\pm$ 0.03} & 4.14 $\pm$ 0.39 & 1.87 $\pm$ 0.56 & \textbf{0.36 $\pm$ 0.08} & 0.95 $\pm$ 0.10 & 5.00 $\pm$ 0.15 & 2.03 $\pm$ 0.20 & 3.45 $\pm$ 0.28 \\
\midrule
\multirow{2}{*}{BA Sim} & PEHNE & 1.48 $\pm$ 0.05 & 3.77 $\pm$ 0.01 & 0.94 $\pm$ 0.27 & \textbf{0.05 $\pm$ 0.01} & 0.68 $\pm$ 0.04 & 3.83 $\pm$ 0.43 & \underline{0.40 $\pm$ 0.11} & 1.32 $\pm$ 0.19 \\
 & CNEE & 0.82 $\pm$ 0.02 & 3.20 $\pm$ 0.01 & 0.95 $\pm$ 0.27 & \textbf{0.05 $\pm$ 0.01} & 0.71 $\pm$ 0.04 & 3.34 $\pm$ 0.51 & \underline{0.39 $\pm$ 0.10} & 1.34 $\pm$ 0.19 \\
\midrule
\multirow{2}{*}{Homophily Sim} & PEHNE & 1.91 $\pm$ 0.15 & 1.29 $\pm$ 0.01 & 0.83 $\pm$ 0.24 & \textbf{0.06 $\pm$ 0.01} & \underline{0.24 $\pm$ 0.03} & 3.75 $\pm$ 0.31 & 0.31 $\pm$ 0.07 & 0.54 $\pm$ 0.14 \\
 & CNEE & 1.52 $\pm$ 0.15 & 0.99 $\pm$ 0.00 & 0.83 $\pm$ 0.24 & \textbf{0.06 $\pm$ 0.01} & \underline{0.25 $\pm$ 0.03} & 3.53 $\pm$ 0.29 & 0.31 $\pm$ 0.07 & 0.57 $\pm$ 0.15 \\
\midrule
\multirow{2}{*}{Coauthor-CS} & PEHNE & 2.16 $\pm$ 0.26 & 2.49 $\pm$ 0.47 & 1.65 $\pm$ 0.33 & \textbf{0.42 $\pm$ 0.08} & 1.33 $\pm$ 0.07 & 3.20 $\pm$ 0.88 & 1.53 $\pm$ 0.35 & \underline{0.90 $\pm$ 0.16} \\
 & CNEE & 1.43 $\pm$ 0.19 & 1.96 $\pm$ 0.38 & 1.37 $\pm$ 0.30 & \textbf{0.38 $\pm$ 0.07} & 1.13 $\pm$ 0.11 & 2.54 $\pm$ 0.76 & 1.24 $\pm$ 0.37 & \underline{0.71 $\pm$ 0.17} \\
\bottomrule
\end{tabular}
\end{adjustbox}
\caption{Test set results (mean $\pm$ SD over five different initializations)
for the \textit{proportion} of treated neighbors used as exposure mapping in the DGP with
the direct influence of $\mathbf{X}_{\mathcal{N}_i}$ on $Y_i$ removed ($\beta_{X_{\mathcal{N}}Y}=0$). Lower is
better; the best-performing method is in bold, the second-best is underlined.}
\label{tab:avg_without}
\end{table}

\begin{table}[t]
\centering
\begin{adjustbox}{width=\textwidth}
\begin{tabular}{llcccccccc}
\toprule
Dataset & Metric & TARNet & NetDeconf & NetEst & TNet & GIN model & SPNet & IDE-Net & HINet (ours) \\
\midrule
\multirow{2}{*}{BC} & PEHNE & \textbf{0.56 $\pm$ 0.20} & 5.12 $\pm$ 0.07 & 2.69 $\pm$ 0.44 & 3.87 $\pm$ 3.56 & \underline{0.70 $\pm$ 0.16} & 6.55 $\pm$ 0.62 & 1.33 $\pm$ 0.33 & 2.24 $\pm$ 0.45 \\
 & CNEE & \textbf{0.54 $\pm$ 0.20} & 4.68 $\pm$ 0.08 & 2.59 $\pm$ 0.44 & 3.87 $\pm$ 3.43 & \underline{0.65 $\pm$ 0.14} & 5.95 $\pm$ 0.57 & 1.31 $\pm$ 0.32 & 2.24 $\pm$ 0.43 \\
\midrule
\multirow{2}{*}{Flickr} & PEHNE & \textbf{0.56 $\pm$ 0.14} & 5.61 $\pm$ 0.20 & 3.70 $\pm$ 0.45 & 0.68 $\pm$ 0.05 & \underline{0.65 $\pm$ 0.08} & 6.70 $\pm$ 0.32 & 3.18 $\pm$ 0.40 & 3.02 $\pm$ 0.50 \\
 & CNEE & \textbf{0.54 $\pm$ 0.12} & 5.18 $\pm$ 0.20 & 3.61 $\pm$ 0.44 & 0.69 $\pm$ 0.05 & \underline{0.61 $\pm$ 0.08} & 6.27 $\pm$ 0.28 & 3.12 $\pm$ 0.44 & 2.99 $\pm$ 0.55 \\
\midrule
\multirow{2}{*}{BA Sim} & PEHNE & 0.65 $\pm$ 0.03 & 4.25 $\pm$ 0.01 & 1.15 $\pm$ 0.11 & 0.67 $\pm$ 0.02 & \textbf{0.50 $\pm$ 0.01} & 4.21 $\pm$ 0.56 & \underline{0.58 $\pm$ 0.10} & 1.09 $\pm$ 0.17 \\
 & CNEE & 0.66 $\pm$ 0.02 & 3.95 $\pm$ 0.01 & 1.14 $\pm$ 0.11 & 0.55 $\pm$ 0.01 & \textbf{0.47 $\pm$ 0.01} & 3.95 $\pm$ 0.53 & \underline{0.54 $\pm$ 0.09} & 1.10 $\pm$ 0.16 \\
\midrule
\multirow{2}{*}{Homophily Sim} & PEHNE & 1.05 $\pm$ 0.02 & 1.19 $\pm$ 0.03 & 0.62 $\pm$ 0.07 & 1.08 $\pm$ 0.02 & \textbf{0.30 $\pm$ 0.03} & 2.17 $\pm$ 0.16 & \underline{0.45 $\pm$ 0.08} & 0.68 $\pm$ 0.11 \\
 & CNEE & 0.82 $\pm$ 0.02 & 0.99 $\pm$ 0.03 & 0.62 $\pm$ 0.07 & 0.88 $\pm$ 0.02 & \textbf{0.29 $\pm$ 0.03} & 1.92 $\pm$ 0.15 & \underline{0.44 $\pm$ 0.08} & 0.69 $\pm$ 0.12 \\
\midrule
\multirow{2}{*}{Coauthor-CS} & PEHNE & 1.29 $\pm$ 0.17 & 2.49 $\pm$ 0.21 & 1.21 $\pm$ 0.03 & \textbf{0.96 $\pm$ 0.06} & 1.78 $\pm$ 0.27 & 3.88 $\pm$ 1.54 & 1.58 $\pm$ 0.28 & \underline{1.12 $\pm$ 0.07} \\
 & CNEE & 1.03 $\pm$ 0.10 & 2.19 $\pm$ 0.31 & 0.89 $\pm$ 0.06 & \textbf{0.77 $\pm$ 0.04} & 1.55 $\pm$ 0.30 & 2.91 $\pm$ 0.84 & 1.20 $\pm$ 0.20 & \underline{0.78 $\pm$ 0.06} \\
\bottomrule
\end{tabular}
\end{adjustbox}
\caption{Test set results (mean $\pm$ SD over five different initializations)
for the (default) \textit{feature-weighted average} of neighbors' treatments used as exposure mapping in the DGP
with the direct influence of $\mathbf{X}_{\mathcal{N}_i}$ on $Y_i$ removed ($\beta_{X_{\mathcal{N}}Y}=0$). Lower is better; the best-performing method is in bold, the second-best
is underlined.}
\label{tab:weight_without}
\end{table}

\begin{table}[t]
\centering
\small
\setlength{\tabcolsep}{6pt}
\begin{tabular}{llcccc}
\toprule
 & & \multicolumn{2}{c}{Proportion (\Cref{tab:avg_without})}
   & \multicolumn{2}{c}{Feature-weighted average (\Cref{tab:weight_without})} \\
\cmidrule(lr){3-4} \cmidrule(lr){5-6}
Dataset & Metric & HINet ($\alpha = 0$) & HINet
                 & HINet ($\alpha = 0$) & HINet \\
\midrule
\multirow{3}{*}{BC}            & PEHNE   & \textbf{0.55 $\pm$ 0.10} & 2.27 $\pm$ 0.28 & \textbf{0.66 $\pm$ 0.11} & 2.24 $\pm$ 0.45 \\
                               & CNEE    & \textbf{0.55 $\pm$ 0.10} & 2.30 $\pm$ 0.29 & \textbf{0.66 $\pm$ 0.11} & 2.24 $\pm$ 0.43 \\
                               & Factual & 0.052 $\pm$ 0.002 & 0.055 $\pm$ 0.001 & 0.054 $\pm$ 0.001 & 0.058 $\pm$ 0.001 \\
\addlinespace[3pt]
\multirow{3}{*}{Flickr}        & PEHNE   & \textbf{0.41 $\pm$ 0.18} & 3.47 $\pm$ 0.27 & \textbf{0.60 $\pm$ 0.09} & 3.02 $\pm$ 0.50 \\
                               & CNEE    & \textbf{0.40 $\pm$ 0.18} & 3.45 $\pm$ 0.28 & \textbf{0.60 $\pm$ 0.10} & 2.99 $\pm$ 0.55 \\
                               & Factual & 0.056 $\pm$ 0.002 & 0.058 $\pm$ 0.003 & 0.092 $\pm$ 0.002 & 0.100 $\pm$ 0.015 \\
\addlinespace[3pt]
\multirow{3}{*}{BA Sim}        & PEHNE   & \textbf{0.60 $\pm$ 0.12} & 1.32 $\pm$ 0.19 & \textbf{0.81 $\pm$ 0.13} & 1.09 $\pm$ 0.17 \\
                               & CNEE    & \textbf{0.61 $\pm$ 0.11} & 1.34 $\pm$ 0.19 & \textbf{0.81 $\pm$ 0.12} & 1.10 $\pm$ 0.16 \\
                               & Factual & 0.054 $\pm$ 0.001 & 0.054 $\pm$ 0.002 & 0.064 $\pm$ 0.002 & 0.064 $\pm$ 0.002 \\
\addlinespace[3pt]
\multirow{3}{*}{Homophily Sim} & PEHNE   & \textbf{0.28 $\pm$ 0.13} & 0.54 $\pm$ 0.14 & \textbf{0.52 $\pm$ 0.13} & 0.68 $\pm$ 0.11 \\
                               & CNEE    & \textbf{0.29 $\pm$ 0.12} & 0.57 $\pm$ 0.15 & \textbf{0.52 $\pm$ 0.13} & 0.69 $\pm$ 0.12 \\
                               & Factual & 0.059 $\pm$ 0.009 & 0.063 $\pm$ 0.008 & 0.070 $\pm$ 0.005 & 0.069 $\pm$ 0.004 \\
\addlinespace[3pt]
\multirow{3}{*}{Coauthor-CS}   & PEHNE   & 0.91 $\pm$ 0.16 & \textbf{0.90 $\pm$ 0.16} & \textbf{0.98 $\pm$ 0.06} & 1.12 $\pm$ 0.07 \\
                               & CNEE    & 0.72 $\pm$ 0.17 & \textbf{0.71 $\pm$ 0.17} & \textbf{0.67 $\pm$ 0.06} & 0.78 $\pm$ 0.06 \\
                               & Factual & 0.331 $\pm$ 0.109 & 0.324 $\pm$ 0.102 & 0.457 $\pm$ 0.096 & 0.473 $\pm$ 0.093 \\
\bottomrule
\end{tabular}
\caption{
Test set results (mean $\pm$ SD over five different initializations) for HINet with $\alpha = 0$ and with $\alpha$ selected by our heuristic, under two DGPs in which the direct influence
of $\mathbf{X}_{\mathcal{N}_i}$ on $Y_i$ is removed by setting $\beta_{X_{\mathcal{N}}Y}=0$. \emph{Factual} denotes the factual validation loss. Lower
is better for all three quantities, and bold marks the better-performing variant
on the counterfactual metrics. For reference, the
irreducible noise variance in the DGP is $\beta_\epsilon^2 = 0.04$, which is the lowest attainable factual loss.}
\label{tab:alpha_scope}
\end{table}
\FloatBarrier
\clearpage
\section{GNN Architectures}\label{app:gnn_arch}

\Cref{tab:gnn_weight,tab:gnn_sum} report test set results for 
HINet with GIN
replaced by GAT \citep{velickovic2017graph}, GraphSAGE
\citep{hamilton2017inductive}, or GCN \citep{kipf2016semi}, under the (default) feature-weighted average mapping 
and the sum mapping, respectively.
Under the feature-weighted average mapping, GraphSAGE generally has the lowest reported error, with GIN second-lowest in every comparison, and the GCN used by prior spillover-effect methods \citep{jiang2022estimating,chendoubly} performing weakest on four out of five datasets. 
Under the sum mapping, GIN
is best on all but one dataset--metric combination. GraphSAGE and GAT degrade sharply, while GCN is far more competitive, sometimes matching
GIN. This is because a sum requires preserving neighborhood size. GIN's sum
aggregation is maximally expressive \citep{xu2018powerful} and preserves the
neighborhood multiplicities and count information. GIN is therefore the only aggregator that is
never far from the best across the different exposure mappings. 
Because the true mechanisms driving interference
are generally unknown, we use GIN throughout for
its consistent performance across mappings, even though a mean-based aggregator can be better when the
underlying dynamics happen to correspond to this form. Our method is modular: any of these GNNs can easily
be substituted without other changes to HINet.

\begin{table}[!ht]
\centering
\footnotesize
\begin{tabular}{llcccc}
\toprule
Dataset & Metric & GIN & GAT & GraphSAGE & GCN \\
\midrule
\multirow{2}{*}{BC} & PEHNE & \underline{0.53 $\pm$ 0.13} & 0.62 $\pm$ 0.19 & \textbf{0.35 $\pm$ 0.10} & 1.10 $\pm$ 0.16 \\
 & CNEE & \underline{0.58 $\pm$ 0.14} & 0.70 $\pm$ 0.19 & \textbf{0.37 $\pm$ 0.13} & 1.17 $\pm$ 0.17 \\
\midrule
\multirow{2}{*}{Flickr} & PEHNE & \underline{0.71 $\pm$ 0.21} & 0.73 $\pm$ 0.13 & \textbf{0.37 $\pm$ 0.08} & 2.20 $\pm$ 0.15 \\
 & CNEE & \underline{0.78 $\pm$ 0.23} & 1.07 $\pm$ 0.14 & \textbf{0.38 $\pm$ 0.08} & 3.79 $\pm$ 0.26 \\
\midrule
\multirow{2}{*}{BA Sim} & PEHNE & \underline{0.28 $\pm$ 0.03} & 0.56 $\pm$ 0.10 & \textbf{0.11 $\pm$ 0.04} & 0.71 $\pm$ 0.14 \\
 & CNEE & \underline{0.35 $\pm$ 0.07} & 0.77 $\pm$ 0.10 & \textbf{0.13 $\pm$ 0.05} & 1.17 $\pm$ 0.14 \\
\midrule
\multirow{2}{*}{Homophily Sim} & PEHNE & \underline{0.17 $\pm$ 0.07} & 0.31 $\pm$ 0.08 & \textbf{0.08 $\pm$ 0.02} & 0.27 $\pm$ 0.06 \\
 & CNEE & \underline{0.20 $\pm$ 0.06} & 0.34 $\pm$ 0.08 & \textbf{0.09 $\pm$ 0.04} & 0.33 $\pm$ 0.05 \\
\midrule
\multirow{2}{*}{Coauthor-CS} & PEHNE & \underline{1.14 $\pm$ 0.23} & \textbf{1.11 $\pm$ 0.06} & 1.20 $\pm$ 0.11 & 1.49 $\pm$ 0.17 \\
 & CNEE & \underline{1.37 $\pm$ 0.34} & 1.62 $\pm$ 0.50 & \textbf{1.32 $\pm$ 0.20} & 1.60 $\pm$ 0.26 \\
\bottomrule
\end{tabular}
\caption{Test set results (mean $\pm$ SD over five different initializations)
for HINet with different GNN architectures, with the (default) \textit{feature-weighted average} of neighbors' treatments used as exposure mapping in the DGP. Lower is better for both metrics.
The best-performing method is in bold; the second-best is underlined.}
\label{tab:gnn_weight}
\end{table}

\begin{table}[!ht]
\footnotesize
\centering
\begin{tabular}{llcccc}
\toprule
Dataset & Metric & GIN & GAT & GraphSAGE & GCN \\
\midrule
\multirow{2}{*}{BC} & PEHNE & \textbf{21.76 $\pm$ 1.57} & 315.86 $\pm$ 51.89 & 446.36 $\pm$ 12.65 & \underline{55.64 $\pm$ 5.18} \\
 & CNEE & \textbf{19.89 $\pm$ 1.52} & 288.56 $\pm$ 39.66 & 396.97 $\pm$ 11.67 & \underline{42.74 $\pm$ 1.49} \\
\midrule
\multirow{2}{*}{Flickr} & PEHNE & \textbf{494.85 $\pm$ 165.23} & \underline{2415.50 $\pm$ 91.02} & 3215.10 $\pm$ 128.85 & 2625.40 $\pm$ 200.76 \\
 & CNEE & \textbf{474.38 $\pm$ 156.37} & 2178.10 $\pm$ 90.63 & 2707.63 $\pm$ 101.83 & \underline{1159.91 $\pm$ 118.60} \\
\midrule
\multirow{2}{*}{BA Sim} & PEHNE & \textbf{10.10 $\pm$ 0.69} & 60.29 $\pm$ 1.38 & 54.99 $\pm$ 0.34 & \underline{13.27 $\pm$ 1.19} \\
 & CNEE & \underline{9.68 $\pm$ 0.62} & 57.32 $\pm$ 0.69 & 54.04 $\pm$ 0.52 & \textbf{8.29 $\pm$ 0.69} \\
\midrule
\multirow{2}{*}{Homophily Sim} & PEHNE & \textbf{0.22 $\pm$ 0.03} & 7.36 $\pm$ 0.10 & 5.78 $\pm$ 0.16 & \underline{2.51 $\pm$ 0.18} \\
 & CNEE & \textbf{0.25 $\pm$ 0.04} & 7.15 $\pm$ 0.07 & 5.91 $\pm$ 0.23 & \underline{2.54 $\pm$ 0.17} \\
\midrule
\multirow{2}{*}{Coauthor-CS} & PEHNE & \textbf{5.68 $\pm$ 0.49} & 152.79 $\pm$ 6.53 & 153.77 $\pm$ 7.35 & \underline{105.68 $\pm$ 8.44} \\
 & CNEE & \textbf{5.33 $\pm$ 0.47} & 147.59 $\pm$ 4.57 & 144.53 $\pm$ 3.58 & \underline{83.73 $\pm$ 8.53} \\
\bottomrule
\end{tabular}
\caption{Test set results (mean $\pm$ SD over five different initializations)
for HINet with different GNN architectures, with the \emph{sum} of neighbors' treatments used as exposure mapping in the DGP. Lower is better for both
metrics. The best-performing method is in bold; the second-best is underlined.}
\label{tab:gnn_sum}
\end{table}


\end{document}